\documentclass{article}

\usepackage[english]{babel}

\usepackage[letterpaper,top=2cm,bottom=2cm,left=3cm,right=3cm,marginparwidth=1.75cm]{geometry}

\usepackage{amsmath}
\usepackage{amsfonts}
\usepackage{graphicx}
\usepackage[colorlinks=true, allcolors=blue]{hyperref}
\usepackage[capitalize,noabbrev]{cleveref}
\usepackage{algorithm}
\usepackage{algorithmic}
\newtheorem{theorem}{Theorem}[section]
\newtheorem{Proposition}[theorem]{Proposition}
\newtheorem{lemma}[theorem]{Lemma}
\newtheorem{Example}[theorem]{Example}

\newtheorem{assumption}[theorem]{Assumption}

\newcommand{\e}{{\mathbf{e}}}
\newcommand{\ie}[0]{\emph{i.e., }}
\newcommand{\VI}{\text{\sc vi}}
\newcommand{\VIL}{\widehat{\VI}^{(\text{\sc lazy})}}
\newcommand{\VIRT}{\widehat{\VI}^{(\text{\sc rt})}}
\newcommand{\VIDR}{\widehat{\VI}^{(\text{\sc dr})}}
\newcommand{\revise}[1]{{#1}}

\usepackage{bm}
\usepackage[title]{appendix}

\newcommand{\amin}{\mathop{\rm arg\min}}

\newcommand{\bbE}{\mathbb{E}}

\newcommand{\bbR}{\mathbb{R}}

\newcommand{\cF}{\mathcal{F}}

\newcommand{\cN}{\mathcal{N}}

\newcommand{\indep}{\perp \!\!\! \perp}
\newcommand{\X}{{\mathbf{X}}}

\newcommand{\bbK}{{\mathbb{K}}}
\newcommand{\bbKj}{\bbK^{(j)}}
\newcommand{\cH}{\mathcal{H}}
\newcommand{\linearh}{\tilde{h}_{\theta_f +\Delta \theta_j}}
\newcommand{\Y}{\mathbf{Y}}
\newcommand{\bfhj}{\mathbf{h}_{\theta_f}^{(j)}}
\newcommand{\bffj}{\mathbf{f}_{0,-j}}
\newcommand{\bmepsj}{\bm{\epsilon}^{(j)}}
\newcommand{\tr}{\text{tr}}
\newcommand{\kerf}{\text{ker}_{\theta_f}}
\newcommand{\ej}{\e^{(j)}}
\renewcommand{\cF}{\mathcal{F}}
\newcommand{\cR}{\mathcal{R}}

\newcommand{\bbP}{\mathbb{P}}
\newcommand{\footremember}[2]{%
    \footnote{#2}
    \newcounter{#1}
    \setcounter{#1}{\value{footnote}}%
}
\newcommand{\footrecall}[1]{%
    \footnotemark[\value{#1}]%
} 

\usepackage[textsize=tiny]{todonotes}
  
\title{Lazy Estimation of Variable Importance for Large Neural Networks
}
 \author{Yue Gao\footremember{uwm}{Department of Statistics,  University of Wisconsin Madison, Madison, WI, USA.}, ~
  Abby Stevens\footremember{uchicagostat}{Department of Statistics,  University of Chicago, Chicago, IL, USA.},~
Rebecca Willet\footrecall{uchicagostat} \footremember{uchicagocs}{Department of Computer Sciences,  University of Chicago, Chicago, IL, USA.}, ~
  Garvesh Raskutti\footrecall{uwm}
}

\begin{document}
\maketitle

\begin{abstract}
As opaque predictive models increasingly impact many areas of modern life, interest in quantifying the importance of a given input variable for making a specific prediction has grown. Recently, there has been a proliferation of model-agnostic methods to measure variable importance (VI) that analyze the difference in predictive power between a full model trained on all variables and a reduced model that excludes the variable(s) of interest. A bottleneck common to these methods is the estimation of the reduced model for each variable (or subset of variables), which is an expensive process that often does not come with theoretical guarantees. In this work, we propose a fast and flexible method for approximating the reduced model with important inferential guarantees. We replace the need for fully retraining a wide neural network by a linearization initialized at the full model parameters. By adding a ridge-like penalty to make the problem convex, we prove that when the ridge penalty parameter is sufficiently large, our method estimates the variable importance measure with an error rate of $O(\frac{1}{\sqrt{n}})$ where $n$ is the number of training samples. We also show that our estimator is asymptotically normal, enabling us to provide confidence bounds for the VI estimates. We demonstrate through simulations that our method is fast and accurate under several data-generating regimes, and we demonstrate its real-world applicability on a seasonal climate forecasting example. 

\end{abstract}

\section{Introduction}
As predictive modeling becomes ubiquitous across a wide swath of application areas, it is especially critical to understand which variables contribute most to making a particular prediction. Black-box machine learning methods are insufficient in the face of algorithmic decision-making about things like sentencing, healthcare, and education, and working toward developing more interpretable methods is becoming more and more relevant \cite{Rudin2019Why,Guidotti_surveybbb}.

Traditional statistical tools based on parametric models (e.g. p-values, ANOVA) for VI inference are dissatisfying due to restrictive assumptions often violated in modern datasets. Non-parametric extensions thus have been explored \cite{doksum1995nonparametric}. In recent decades, many VI methods designed for modern deep learning models have been investigated; most of these methods are gradient-based and depend on the structure and the weights of nodes in a given specific neural network \cite{shrikumar_learning_2019, sundararajan2017axiomatic, smilkov2017smoothgrad, bach2015pixel}. Few statistically rigorous properties are provided for these methods, and the VI definition is always intimately attached to the network itself, making it hard to interpret in a model-agnostic setting.

In a model-agnostic setting, a natural definition of VI that is independent of the estimation procedure is to measure the loss of predictive power when the variables of interest are deleted. To estimate such model-agnostic VI, \emph{retraining} is the most widely used type of method, which involves training separate models on the reduced data with the variables of interest deleted and assessing the predictive skill difference \cite{williamson_unified_2020, jinglei2018, sapp2014targeted}. Retraining often acts as the best benchmark to evaluate other VI estimation methods \cite{hooker_benchmark_2019} due to its accuracy, yet it is computationally infeasible in high-dimensional settings.
Other methods for VI estimation include knockoff methods \cite{barber2018knockoff, candes_panning_2017} and Floodgate \cite{zhang_floodgate_2021}, which  require the co-variate distribution to be known. 
An alternative approach is to use a \emph{dropout}-type method \cite{chang2017dropout}. Dropout is best-suited for assessing how much a variable affects a predictive model, as opposed to our goal of assessing how much a variable affects the response. Despite the resulting issues with VI estimation accuracy, it is still widely used  in practice as a proxy for VI due to its computational tractability. 

In this work, we propose a computationally efficient variable importance estimation procedure for model-agnostic and distribution-free settings with theoretical guarantees that leverages a \textit{lazy retraining} framework inspired by \cite{chizat_lazy_2020}. 
The key idea is to train a new model on the transformed training data, akin to retraining, but on a linearized version of the model centered around model parameters learned from the original (unreduced) training data. We perform ridge regression on this linearized model in the gradient feature space, meaning that our lazy retraining procedure can be computed very quickly. The resulting method, when applied to wide neural network models, admits error bounds that show it is nearly as accurate as full retraining, while computationally it is nearly as fast as dropout. Our theoretical bounds are complemented by a collection of simulations that explore the limitations of dropout and benefits of lazy retraining under a variety of conditions and an application to understanding the importance of various climate indices in a seasonal forecasting task.

In summary, the main contribution of this paper is a \textbf{new, computationally efficient VI estimation method with statistical performance guarantees} in a model-agnostic and distribution-free setting when using large neural networks. Our theoretical  analysis facilitates statistical inference, and we illustrate our approach on both synthetic and real-world data to support the theoretical claims and demonstrate the utility of our method. Other empirically-driven VI estimation methods exhibit similarities to our approach; our theoretical analyses may provide new insights into those methods as well as the one we propose in this paper.

\section{Notation and Preliminaries}
Suppose we have samples $Z_i = (\X_i, Y_i), i=1,\dots, n$ for data $Z = (X,Y) \sim P_0$, where $\X_i\in \bbR^p$ is the $i$-th  $p$-dimensional feature vector and $Y_i$ is the $i$-th observed response. $X$ denotes the multi-variate random variable containing features, $Y$ denotes the response random variable.
Let $X_{-j}\in \bbR^{p-1}$ (resp. $\X_{i,-j}$) denote the features in $X$ (resp. $\X_{i}$) with the $j$-th variable removed; on the other hand, if we replace the $j$-th random variable in $X$ (resp. $\X_{i}$) by its marginal mean $\mu_j = \mathbb{E}(X_j)$, 
we denote it as $X^{(j)}$ (resp. $\X_{i}^{(j)}$), \ie 
$X^{(j)} = (X_1, \dots, X_{j-1},\mu_j, X_{j+1}, \dots, X_{p})$.
Let $P_{0}, P_{0,-j}$ be the population distributions for $X$ and $X_{-j}$ 
and let $P_n, P_{n,-j}$ be the empirical distributions of $X$ and $X_{-j}$ for $j \in [p]$. 
$\delta_{Z_i}$ denotes the point mass probability measure at the $i$-th observation $Z_i$. \revise{We denote $\bbE_0, \bbE_{0,-j}$ as the expectations taken with respect to $P_0$ and $P_{0,-j}$.}

Let $f_{0}$ denote the true function mapping $X$ to the expected value of $Y$ conditional on $X$, and let $f_{0,-j}$ denote the function mapping $X^{(j)}$ to the expected value of $Y$ conditional on $X^{(j)}$:
\begin{align}
    f_{0}(X) :=& \bbE_{0}[Y|X]; \\
     f_{0,-j}(X^{(j)}) :=& \bbE_{0,-j}[Y|X_{-j}] .
\end{align}

Let $f_n$ be the empirical model trained using all $p$ variables in $X$ within a certain function class $\mathcal{F}$(we refer to this as the \textit{full model}):
\begin{equation}
f_n \in \amin_{f\in\mathcal{F}}\frac{1}{n} \sum_{i=1}^n [Y_i - f(\X_i)]^2.
\end{equation}

To measure the accuracy of an approximation $f_n(x)$ to its target function $f_0$, we use the $L_2(\mu)$-norm
\begin{equation}
    \|f_n - f\|^2 = \int | f_n(x)-f(x)|^2 d \mu(x),
\end{equation}
where $\mu$ is the probability measure for $X$.

Further, we use $\epsilon$ and $\epsilon^{(j)}$ to denote the respective remainder terms for any $j \in [p]$:
\begin{equation}
    \epsilon := Y - \bbE_0[Y|X]; ~~ \epsilon^{(j)} := Y  - \bbE_{0,-j}[Y|X_{-j}].
    \label{eq: remaining terms}
\end{equation}

We will define our measure of variable importance (VI) in terms of a \textit{predictive skill measure} $V(f,P)$ (the same measure in \cite{williamson_unified_2020}). Larger values of $V(f,P)$ should indicate better predictive performance. For $Z = (X, Y)$, we denote $\dot{V}(f, P; \delta P)$ as the Gateaux derivative of $V(f,P)$ at $P$ in the direction $\delta P$.
Specifically, one of the predictive skill measures we consider is the negative mean squared error (MSE): 
\begin{equation}
    V(f, P) = - \bbE_{(X, Y)\sim P} [Y - f(X)]^2,
    \label{eq: negative MSE}
\end{equation}
and the corresponding $\dot{V}(f, P; \delta P)$ is $\dot{V}(f, P; \delta P) = -\int_{Z = (X,Y)} (Y - f(X))^2 d(\delta P)$.
Hence, the Gateaux derivative of the negative MSE is
$\dot{V}(f, P_0; \delta_{Z_i} - P_0) = - (Y_i - f(X_i))^2 + \bbE[Y - f(X)]^2
$
and
$\bbE[\dot{V}(f_0, P_0; \delta_{Z_i} - P_0)] = 0$.

\section{Estimating Variable Importance}
The VI measure we consider, which makes no assumptions on the data generating mechanism, is 
\begin{equation}
    {\VI_j} := V(f_0,P_0) - V(f_{0,-j}, P_{0,-j}).
        \label{eq:VIdef}
\end{equation}

$\VI_j$ quantifies the difference in predictive skill between the full model and the reduced model for any $j \in [p]$. Consider the following simple linear model example, where we take the negative MSE as the predictiveness  measure.

\begin{Example}\label{ex:linear}
    Suppose $Y = \beta_1 X_1 +\beta_2 X_2 +\epsilon$, where $X_i\sim \mathcal{N}(0,\sigma^2),i = 1,2, ~ \text{Cov}(X_1, X_2) = \rho$,
    and $\epsilon$ is a $\cN(0, \sigma_{\epsilon}^2)$ noise that is independent of the features.
    The variable importance of the first variable is
        $$\VI_1 = \beta_1^2\cdot \text{Var}(X_1|X_2) = \beta_1^2(1-\rho^2)\sigma^2$$
    due to the fact that $X_1|X_2 \sim \cN(\rho X_2, (1-\rho^2)\sigma^2)$ (see \Cref{proof: example})
\end{Example}

In general, we see from this example that the variable importance measure is determined not only by the relationship between $X_j$ and $Y$, but also the covariance structure in the features.

Our goal is to estimate $\VI_j$ for any variable $X_j$ from data $\{(\X_i, Y_i) \}_{i=1}^n$ with no assumptions on the relationship between $X$ and $Y$. 
For empirical estimators $f_n$ and $f_{n,-j}$ of $f_0$ and $f_{0,-j}$, a plug-in estimator of our VI measure is

\begin{equation}
\widehat{\VI}_j = V(f_n, P_n) - V(f_{n,-j}, P_{n,-j}).
\end{equation}
The key problem we are concerned with in this paper is how to estimate $f_{n,-j}$ in an accurate and computationally efficient way. Traditionally, people use the following two types of methods to do the estimation: \textit{dropout} and \textit{retraining}. 

\subsection{Dropout}

The method we are calling \textit{dropout} estimates $\mathbb{E}(Y|X_{-j})$ by plugging the dropout features $X^{(j)}$ into the full model $f_n$. 
In this case, the variable importance measure can be estimated by
\begin{equation}
    \VIDR_j = V(f_n, P_n) - V(f_n, P_{n,-j}).
\end{equation}
For the negative MSE measure of predictive skill for instance, 
the dropout estimate measures the difference between the squared error on the \textit{original} training set  and the squared error on the  training set \textit{after replacing feature $j$ with its mean}.
Dropout is superior among all plug-in estimators in terms of computational cost -- we only need to train the model once to get $f_n$. This is desirable, especially when the function class $\mathcal{F}$ is large and complicated, such as  with neural networks, and the computational cost for training the model is high. Despite this benefit, dropout is unreliable in many settings, as we will revisit in \Cref{subsection: comparison}.

\subsection{Retrain}
An alternative to dropout is what we call \textit{retraining}. Given a function class $\mathcal{F}$, the retraining method estimates $\VI_j$ by training separate models 
\begin{equation}{f}_{n,-j} \in \amin_{f\in \mathcal{F}} [Y_i - f(\X_i^{(j)})]^2\end{equation}
for each variable $j\in [p]$ to estimate $f_{0,-j}$. 
Hence, VI under this framework is estimated via
\begin{equation}
    \VIRT_j = V(f_n, P_n) - V(f_{n,-j}, P_{n,-j}).
\end{equation}
When taking negative MSE as the predictive skill measure, the retraining estimate in this case measures the difference between the squared error of a model trained \textit{without} feature $j$ and the squared error of a model trained \textit{with} feature $j$.
Retraining  is more accurate than dropout as long as the function class $\mathcal{F}$ is large enough, but requires training $p+1$ models, which can be prohibitively computationally expensive in many settings. In this paper, we are especially interested in the setting when the function class is as large as a wide neural network.

\subsection{Dropout vs. Retrain for Linear Models}\label{subsection: comparison}

The dropout method is widely used to estimate variable importance due to its efficiency. However, in cases where variables in $X$ are highly correlated, dropout behaves problematically. Below, we will illustrate and quantify the difference of the variable importance estimation in the random design linear model case, where we take the negative MSE as the $V(f,P)$ measure as in \cref{eq: negative MSE}. For simplicity, we restrict the function space $\mathcal{F}$ to the linear function space here.

Suppose $X\in \bbR^p\sim \mathcal{N}(0, \Sigma)$, $\epsilon \sim \mathcal{N}(0,\sigma_{\epsilon}^2)$. Assume $\Sigma$ is positive definite.
Let $\beta^* := \amin_{w\in \bbR^p} \bbE [Y - X^{\top}w]^2$, so $\beta^* = \Sigma^{-1} \bbE(XY)$. 
In the population version, the dropout method uses the predictor $X_{-j}^{\top} \beta^*_{-j}$ 
(where $\beta^*_{-j} \in \bbR^{p-1}$ is $\beta^*$ with its $j$-th element removed)
to estimate $\bbE(Y|X_{-j})$,
while the retraining method uses the predictor $X_{-j}^{\top}{\beta}^{(j)}$, where ${\beta}^{(j)} \in \bbR^{p-1}$ is ${\beta}^{(j)} = \amin_{w\in \bbR^{p-1}} \bbE[Y - X_{-j}^{\top}w]^2$.
The following proposition characterizes the difference between VI estimates corresponding to the retraining and dropout methods.
\begin{Proposition}\label{prop:rtdr}
    In the linear function space, the difference between the variable importance estimates for variable $j$ from the population version of the dropout and retraining methods is:
    \begin{equation*}
        \begin{split}
            &\VIDR_j- \VIRT_j \\=& \frac{\vec{\boldsymbol{\gamma}_j}^{\top} \Sigma_{(j)}^{-1}\vec{\boldsymbol{\gamma}_j}}{(\Sigma_{jj} - \vec{\boldsymbol{\gamma}_j}^{\top} \Sigma_{(j)}^{-1}\vec{\boldsymbol{\gamma}_j})^2} \left[
                \mathbb{E}(X_j Y) - \vec{\boldsymbol{\gamma}_j}^{\top} \Sigma_{(j)}^{-1}
                \mathbb{E}(X_{-j}Y)
            \right]^2,
        \end{split}
    \end{equation*}
    where $\vec{\boldsymbol{\gamma}_j} = \mathbb{E}(X_j X_{-j})\in \mathbb{R}^{p-1} $.
\end{Proposition}

If the true model between $Y$ and $X$ is linear, \ie $Y = X^{\top}\beta^* +\epsilon$, and $X \indep \epsilon$, the variable importance estimated by retraining linear regression is:
\begin{equation}
        \VIRT_j = {\beta_j^*}^2 (\Sigma_{jj} - \vec{\boldsymbol{\gamma}_j}^{\top} \Sigma_{(j)}^{-1} \vec{\boldsymbol{\gamma}_j});
\end{equation}
furthermore, in this setting $\VIRT_j$ is exactly the true variable importance defined in \eqref{eq:VIdef}.
In contrast, the dropout framework will give
\begin{equation}
        \VIDR_j = {\beta_j^*}^2 \cdot \Sigma_{jj} .
\end{equation}
If feature $j$ is important and highly correlated with feature $k$ (but independent of all other features), then $\vec{\boldsymbol{\gamma}_j}^{\top} \Sigma_{(j)}^{-1} \vec{\boldsymbol{\gamma}_j}$ may be very large, making the difference between $\VIDR_j$ and $\VIRT_j$ similarly large. This example illustrates how dropout can significantly overestimate variable importance, even in simple settings.

\section{Lazy Training}
Our central interest is in inferring VI  using complex models that are time-consuming to train, making the baseline retraining method described above computationally infeasible. With this in mind, we turn our attention to neural network (NN) models, a setting in which dropout is widely used.

Motivated by the need for faster and more accurate methods for estimating VI with NN, we propose a computationally efficient VI estimate inspired by the lazy training framework of \cite{chizat_lazy_2020} that estimates the difference between the full model parameters and the model parameters  when the $j$-th variable is removed. Like dropout, our procedure only requires us to train the NN once on the full data, and then we solve a linear system to update the full model parameters for each variable $j \in [p]$.

Given the training data $\{(\X_i^{(j)}, Y_i)\}$ sampled from $(X,Y) \sim P_0$ for $i = 1, \dots, n$ and the underlying function $f_{0}(X) = \bbE_{P_0} [Y|X]$,
there exists a 
a neural network function class $\{h_{\theta}(x): \bbR^p \mapsto \bbR| \theta\in \bbR^M\}$ that is parameterized by a vector $\theta$, such that when we train the model parameters over this class by 
\begin{equation}
    \theta_f = \amin_{\theta\in \bbR^M} \frac{1}{n}\sum_{i=1}^{n} [Y_i - h_{\theta}(\X_i)]^2,
\end{equation}
the estimation error can be bounded by
 $\|h_{\theta_f}(x) - f_0(x) \| = O(n^{-1/2})$ up to some log terms \cite{barron1994approximation}. To achieve this, the scale of the number of parameters $M$ depends on the complexity of the target function. 

For very complex functions, we can still achieve this accuracy with $M =O(\sqrt{n})$.
 
In order to estimate $\VI_j$, we need an estimate of what we are calling the \textit{reduced} model $h_{\theta_{-j}}$, where
\begin{align}
   \theta_{-j} = \amin_{\theta\in \bbR^M} \frac{1}{n}\sum_{i=1}^{n} [Y_i - h_{\theta}(\X_i^{(j)})]^2.
\end{align}

Instead of retraining a NN to estimate $\theta_{-j}$, we can instead estimate the difference between the full model parameters $\theta_f$ and $\theta_{-j}$ using this linear approximation, and simply update the full model parameters with this correction to estimate $h_{\theta_{-j}}$. We are essentially regressing the error resulting from the dropout estimation against the gradient to estimate this correction, and to do so we solve the following convex problem based on the training data $\{(\X_i^{(j)}, Y_i)\}$ for $i = 1, \dots, n$ and a $2$-norm penalty on the parameters:
\begin{align}
       \Delta \theta_j (\lambda, n)
    = \amin_{\omega\in \bbR^{M}}&\Big\{ \frac{1}{n}\sum_{i = 1}^{n} \big[Y_i - h_{\theta_f}(\X_{i}^{(j)}) 
    \label{eq: delta_theta}
    \\
    & -  \omega^\top \nabla_{\theta} h_{\theta}(\X_{i}^{(j)}) |_{\theta = \theta_f}\big]^2 + \lambda \|\omega\|_2^2 \Big\}, \nonumber
\end{align}
where $\lambda>0$ is the penalty parameter. 

Accordingly, the reduced neural network parameters are $ \Delta \theta_j(\lambda,n) + \theta_{f}$. 
For the simplicity of notation, we write $\Delta \theta_j(\lambda,n)$
as $\Delta\theta_j$ for short.
Then the reduced model approximation without the $j$-th feature is $\bbR^p\mapsto \bbR: x \mapsto h_{\theta_f + \Delta\theta_{j}}(x)$.

Hence, the variable importance measure under lazy training is
\begin{equation}
    \VIL_j = V(h_{\theta_f}, P_n) - V(h_{\theta_f+\Delta \theta_j}, P_{n,-j}).
\end{equation}
Under the negative MSE measure $V(f,P)$, we have
\begin{equation*}
    \VIL_j  = \frac{1}{n}\sum_{i=1}^n \{[Y_i - h_{\theta_f+\Delta \theta_j}(\X_i^{(j)})]^2 - [Y_i - h_{\theta_f}(\X_i)]^2\}.
\end{equation*}
(More precisely, we use data splitting for training and estimating VI as detailed in \cref{alg:cap}.)
Essentially, the linearized approximation of the NN is linear in the gradient feature map $x\mapsto \nabla_{\theta} h_{\theta}(x)|_{\theta_f}$. In fact, this gradient feature map induces the Neural Tangent Kernel (NTK, \cite{jacot2020neural}): for any $x, x'\in \bbR^p$,
\begin{equation}
    \text{ker}_{\theta_f}(x, x') := \langle \nabla_{\theta} h_{\theta}(x)|_{\theta_f},\nabla_{\theta} h_{\theta}(x')|_{\theta_f}\rangle.
\end{equation}
Thus $\Delta \theta_j$ can be viewed as the solution for a kernel ridge regression problem with  kernel $\text{ker}_{\theta_f}$.

\subsection{Theoretical Guarantee}

By \cite{williamson_unified_2020}, when the empirical estimates for 
 $\bbE(Y|\X)$ and $\bbE(Y|\X_{-j})$ converge to the target functions $f_0$ and $f_{0,-j}$ at the rate of $O_p(n^{-1/4})$ in function norm, we achieve an asymptotically normal and efficient estimator for the VI measure. In this section, we  give a theoretical guarantee to show that the lazy prediction  $h_{\theta_f + \Delta \theta_j}(X^{(j)})$ for the reduced model achieves such convergence rate, so that the lazy training procedure gives an accurate estimate of VI with an error in the order of $O(\frac{1}{\sqrt{n}})$ and we can make inference accordingly.

Let $\e^{(j)}$ denote the difference between the true reduced function $f_{0,-j}(X^{(j)})$ and the corresponding dropout estimation:
\begin{equation}
        \e^{(j)} := f_{0,-j}(\X^{(j)}) - h_{\theta_f}(\X^{(j)}) \in \bbR^n.
        \label{def: e^j}
\end{equation}
Further, we denote the kernel matrix on $X^{(j)}$ induced by the gradient feature map as $\bbK^{(j)}\in \bbR^{n\times n}$, whose elements are defined as:
\begin{equation}
    \bbK^{(j)}_{ik} := \text{ker}_{\theta_f}(\X_i^{(j)}, \X_k^{(j)}),~ i,k\in [n].
\end{equation}

\revise{Before diving into the main results, we first clarify two types of notation for order of approximation:
\begin{itemize}
    \item 
    $f(n) = O(g(n))$ if there exists $M>0$ and $N>0$, such that 
    $|f(n)| \leq M g(n)$ for all $n>N$.
    \item $X_n = O_p(a_n)$ as $n\rightarrow \infty$ if for any $\epsilon>0$, there exists $M>0$ and $N>0$, such that $\bbP(|\frac{X_n}{a_n}| >M)< \epsilon$ for any $n>N$.
\end{itemize}
}
\begin{assumption}
\label{asspt: kernel bound}
     For any $j\in [p]$ \revise{and the regularization parameter $\lambda = O(\sqrt{n})$, we assume:
     \begin{itemize}
         \item[(a)]  $ \|[\bbK^{(j)} + \lambda I_n]^{-1}\e^{(j)}\|^2 = O_p(1/\sqrt{n})$;
         \item[(b)]\label{asspt: trace} $\text{tr}(\bbK^{(j)}) = O_p(n)$.
     \end{itemize}
      }
     
\end{assumption}

The above assumption (b) is commonly used in NTK literature (see e.g. \cite{hu2019simple}). \revise{For a two-layer neural network, we can verify this numerically (see \Cref{appendix: trace_div}). For the assumption (a), by the fact that $\bbK^{(j)}$ is positive semi-definite, this assumption can be satisfied when we have a large regularization $\lambda = O(\sqrt{n})$.}

\begin{assumption}
\label{asspt: tail bound}
    For the noise term $\epsilon^{(j)}$, we have the following assumption on its conditional tail probability: there exists $\sigma$ such that for any $j \in [p]$,
\begin{equation}
    \bbE\left[e^{\lambda \epsilon^{(j)}}| X^{(j)} \right] \leq e^{\sigma^2\lambda^2/2}, ~\text{ for all }\lambda \in \bbR.
\end{equation}
\end{assumption}

\begin{assumption}
\label{asspt: correlated}
    Denote the gradient feature matrix as  $\Phi\in \bbR^{n\times M} = (\nabla_{\theta} h_{\theta}(\X_1)|_{\theta = \theta_f},\dots, \nabla_{\theta} h_{\theta}(\X_n)|_{\theta = \theta_f})^{\top}$.
    We assume $\| \Phi^{\top} \e^{(j)} \|_2 \leq O_p(1)$.
\end{assumption}

This assumption essentially requires that the linear space of neural tangent kernels can well represent $\e^{(j)}$. We know that $\e_i^{(j)}$ is a function of $\X_i^{(i)}$, thus as long as the neural network function class is large enough, this can be satisfied with respect to the sample size $n$.

\begin{theorem}\label{thm1}
    Suppose \Cref{asspt: kernel bound}, \ref{asspt: tail bound} and \ref{asspt: correlated} hold, then for a neural network structure $h_{\theta}(\cdot)$ which is $L$-smooth with respect to its parameters $\theta$, as long as we take the ridge penalty parameter in the order $\lambda = O(n^{1/2})$, then the lazy training method can accurately predict the reduced model without the $j$-th covariate, \ie
    \begin{equation}
        \|h_{\theta_f + \Delta \theta_j}(x) - \bbE(Y| X^{(j)})\|_{2} = O_p(n^{-1/4}).
    \end{equation}
    Therefore our variable importance estimator $\VIL_j$
    is asymptotically normal and has an error rate $O_p(n^{-1/2})$:
    \begin{equation}
        \VIL_j  - \VI_j = \Delta_{n,j} + O_p(n^{-1/2});
    \end{equation}
    where 
    \begin{align}\label{eq:normal}
        \Delta_{n,j} & = \frac{1}{n} \sum_{i = 1}^n
        \big[\dot{V}(f_0, P_0;\delta_{Z_i}-P_0)  \\
        &-\dot{V}(f_{0,-j}, P_{0,-j};\delta_{Z_i}-P_{0,-j})\big]
        \rightarrow_d \cN(0, \tau_{n,j}^2); \nonumber
    \end{align}
    here the variance is $\tau_{n,j}^2 = \text{Var}({\epsilon^{(j)}}^2 - \epsilon^2)/n$, where $\epsilon$ and $\epsilon^{(j)}$ is defined in \cref{eq: remaining terms}.
\end{theorem}

This result enables us to construct Wald-type confidence intervals around our LazyVI estimates. In particular, the $\alpha-$level confidence intervals are given by
\begin{equation}\label{ci}
    \VIL_j  \pm z_{\frac{\alpha}{2}} \hat{\tau}_{n,j}
\end{equation}
where $\hat{\tau}_{n,j}$ is the plug-in estimate of $\tau_{n,j}$ in \eqref{eq:normal} and $z_{\frac{\alpha}{2}}$ is the $\alpha/2$ quantile of the standard normal distribution. 

\subsection{Proof Overview}
The challenge of proving  Theorem \ref{thm1} is to bound the error of the lazy neural network trained using data without a certain variable -- note that we are bounding the estimation error ($\|h_{\theta_f + \Delta \theta_j} - f_{0,-j}\|$) instead of the prediction error ($\|h_{\theta_f + \Delta \theta_j} - f_{0}\|$) that is the focus of much of the deep learning community, since the predictive 
skill
of the reduced model is expected to decrease when an important variable is removed. At a high level, our proof reduces the estimation error of the neural network from lazy training to the error between the NTK estimation and the target function, where we use techniques from kernel ridge regression. The difference here is that most NTK papers (see e.g. \cite{jacot2020neural}) use random initialization for the parameters and optimization without penalty, while our method starts from a specific initialization (the full model), and requires the penalty parameter $\lambda$ to be large ($\lambda = O(n^{1/2})$) to ensure convergence.

The following two lemmas give some intuition on how the neural network trained by the lazy procedure can accurately estimate the reduced model. Basically, the bound for the error consists of two parts: the error from the kernel ridge regression (discussed in \Cref{lemma: KRR}), and the error from the linear approximation of the neural network (in \Cref{lemma: linear_approx}). More proof details are deferred to the Appendix. 

Denote the linear approximation of the  network as
\begin{equation}\label{eq: linearization}
    \tilde{h}_{\theta_f + \Delta \theta_{j}}(x) := h_{\theta_f}(x) + \langle \nabla_{\theta} h_{\theta}(x) |_{\theta = \theta_f}, \Delta \theta_{j}\rangle.
\end{equation}

\begin{lemma}
\label{lemma: KRR}
Let $\lambda$ be penalty parameter in \Cref{eq: delta_theta}, we have with probability at least $1-\delta$,
\begin{align}
        &\| \tilde{h}_{\theta_f + \Delta \theta_{j}}(X^{(j)}) - f_{0,-j} (X^{(j)})\|_n \\
        &\leq  \frac{\lambda \|[\bbK^{(j)} +\lambda I_n]^{-1}\e^{(j)}\|}{\sqrt{n}} + \sigma\sqrt{\frac{ \text{tr}[\bbK^{(j)}]} {4\lambda n}} + 
        \sigma \sqrt{\frac{2\log(1/\delta)}{n}}. \nonumber
\end{align}
\end{lemma}

\Cref{lemma: KRR}
combined with \Cref{asspt: kernel bound} when the penalty parameter is $\lambda = O(n^{1/2})$, yields a bound on the empirical error of the kernel ridge regression component of  $O_p(n^{-1/4})$. Based on this empirical bound, we could then further bound the generalization error of the estimated function using function complexity (See \Cref{appendix: generalization_error}).

\begin{lemma}
\label{lemma: linear_approx}
For a large neural network with width in the order $O(\sqrt{n})$, with high probability we have for all $j \in [p]$,
\begin{equation}
    \|  \linearh(x) - h_{\theta_f + \Delta \theta_j}(x)\| \leq O(n^{-1/4}).
\end{equation}
\end{lemma}
 \Cref{lemma: linear_approx} shows that as long as the neural network is sufficiently large, the neural network with updated parameters $\theta_f + \Delta_j$ \revise{is close to its linear approximation}.

\subsection{Implementation}
We estimate $h_{\theta_f}$ and $h_{\theta_{-j}}$ using $n_1 < n$ samples as training data, and use the remaining $n_2 = n - n_1$ samples to estimate VI. For the dropout method, VI is estimated simply by plugging the modified testing data $\{\X_{i}^{(j)}\}_{i=n_1+1}^{n}$ into $h_{\theta_f}$. For the retraining method, first $h_{\theta_{-j}}$ is estimated by retraining the NN $h$ with $\{\X_{i}^{(j)}\}_{i=1}^{n_1}$, and then VI is estimated by plugging the modified testing data into this retrained estimate.

For the lazy training method, which we call LazyVI, we use the training data to estimate the full model parameters, compute the gradient of the network with respect to each model parameter for each training sample, and then regress these gradients against the difference between $Y$ the dropout estimates from the training data to estimate the parameter correction $\Delta  \theta_j$ for variable $j$. We then update the full model parameters using this learned correction to compute the VI estimate and its associated standard errors. See Algorithm \ref{alg:cap} for full details.

Theorem \ref{thm1} makes the assumption that the ridge parameter $\lambda$ from Equation (\ref{eq: delta_theta}) is large. Since we are ultimately interested in estimating $h_{\theta_{-j}}$ and not $\Delta \theta_j$, we evaluate $h_{\theta_f + \Delta \theta_j}(\cdot)$ through K-fold CV to choose $\hat{\lambda}_j$ for each variable (\Cref{appendix:ridge_param} in \cref{appx:cv_alg}). \revise{Our implementation is available at \url{https://github.com/Willett-Group/lazyvi}.}

\begin{algorithm}[tb]
    \caption{Lazy training for VI}
    \label{alg:cap}
\begin{algorithmic}
\REQUIRE Data: $\{\X_i, Y_i\}_{i=1}^n$; $\lambda > 0$; training size: $0<n_1<n$; $n_2 \gets n-n_1$; NN structure: $\theta\in\bbR^M \mapsto h_{\theta}(\cdot)$
{LazyVI}{$\{\X_i, Y_i\}_{i=1}^n$; $\lambda$, $n_1$}
\STATE $\theta_f \gets \amin_{\theta\in \bbR^M} \frac{1}{n_1} \sum_{i=1}^{n_1} [Y_i - h_{\theta}(\X_i)]^2 $
\STATE $v_n \gets -\frac{1}{n_2}\sum_{i=n_1+1}^{n}[Y_i - h_{\theta_f}(\X_i)]^2 $
\FOR{$j \in [p]$}
    \STATE $\X_i^{(j)} \gets \X_i$; $\X_{ij}^{(j)} \gets \frac{1}{n_1}\sum_{i=1}^{n_1}\X_{ij}$
    \STATE $\e_i^{(j)} \gets Y_i - h_{\theta_f}(\X_{i}^{(j)}), ~i=1,\dots,n_1$
    \STATE $\Phi_i^{(j)} \gets \nabla_{\theta} h_{\theta}(\X_{i}^{(j)}) |_{\theta = \theta_f},~i=1,\dots,n_1$
    \STATE $\Delta \theta_j \gets \displaystyle{\amin_{\omega\in \bbR^{M}}} \frac{1}{n_1}\sum_{i = 1}^{n_1} 
        [\e_i^{(j)} -  
        \omega^\top \Phi_i^{(j)}
        ]^2  + \lambda \|\omega\|_2^2$
    \STATE $v_{n,-j} \gets \displaystyle -\frac{1}{n_2} \sum_{i=n_1+1}^{n}[Y_i - h_{\theta_f+\Delta \theta_j}(\X_i^{(j)})]^2$
    \STATE $\widehat{\VI}_j \gets v_{n}-v_{n,-j}$
    \STATE $t_{i,j} \gets (Y_i -  h_{\theta_f+\Delta \theta_j}(\X_i^{(j)}))^2 -(Y_i -  h_{\theta_f}(\X_i))^2$
    \STATE $\hat{\tau}_j \gets \frac{1}{n_2}\sum_{i=1}^{n_2} (t_{i,j} - \bar{t}_j)^2/{n_2}$
\ENDFOR

\ENSURE $\widehat{\VI}_j, ~j=1,\dots,p$.
\end{algorithmic}
\end{algorithm}

\section{Simulations}
We first assess the performance of LazyVI on simulated data to highlight key theoretical claims and assumptions and show that our method is empirically practical.  For these experiments, we train a wide, fully connected two-layer neural network with ReLU activation for all simulations. Unless otherwise specified, the width of the hidden layer in the training network is $m=50$.

\subsection{Impact of Correlation in Linear Systems}
Our first set of simulations serve to support key details of our theoretical analysis. We consider data generated from the linear model $f(X) = 1.5X_1 + 1.2X_2 + X_3 + \epsilon$, where $\epsilon \sim \cN(0, 0.1)$ and $X \sim \cN(0, \Sigma_{6 \times 6})$, so the response only depends on the first three of the six variables. All variables are independent except for $X_1$ and $X_2$, whose correlation is $\rho$. As discussed in Example \ref{ex:linear}, the true VI of $X_1$, $X_2$, and $X_3$ are given by $(1.5)^2 (1-\rho^2)$, $(1.2)^2(1-\rho^2)$, and $1$, respectively, and the VI of the remaining 3 variables is zero. In this simple setting, we find that LazyVI approximates the true $\VI$ well with desirable coverage and a considerable speed-up relative to retraining (\Cref{appendix:linear_exp}).

\begin{figure}[ht]
\centerline{\includegraphics[scale=0.16]{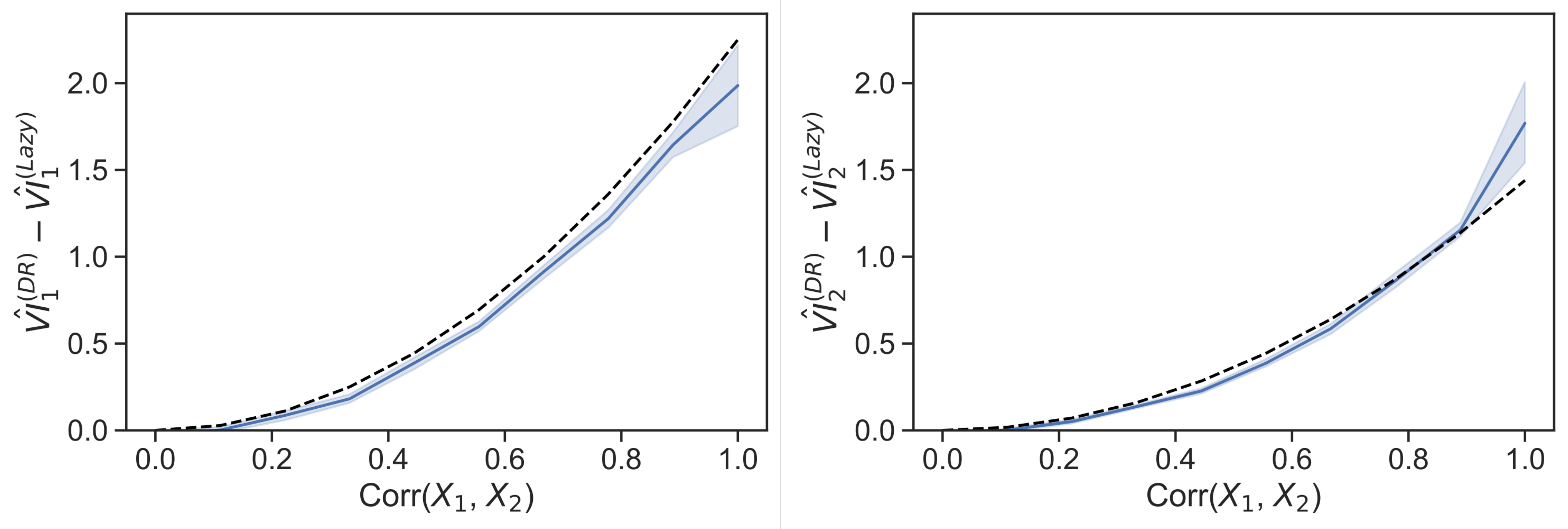}}
\caption{\label{fig:corr} Difference between the dropout and LazyVI estimates for $X_1$ and $X_2$. Dotted line is theoretical gap and shading shows std. across 10 repetitions. We see that LazyVI closely approximates the theoretical variable importance measures under increasing correlations.}
\end{figure}

We show in Prop.~\ref{prop:rtdr} that, when data are generated from a linear model, the difference between the dropout and retraining variable importance estimates is a function of the covariance of $X$. After training the full model, we use both the dropout and our lazy procedure to estimate VI for increasing values of $\rho$. In Figure \ref{fig:corr}, we show the difference between the dropout and LazyVI estimates for variables $X_1$ and $X_2$ alongside the analytic difference between $\VIDR$ and $\VI$ (dotted line). We see that the gap between LazyVI and dropout evolves with $\rho$ according to the theoretical analysis, providing evidence that LazyVI behaves as expected. 

We use this simple linear setting to explore two additional assumptions from our theoretical results. First, the linearization in  \eqref{eq: linearization} is a first order Taylor approximation and assumes the full model parameters are close to the reduced model parameters. If we try to linearize a neural network around a random initialization, our LazyVI estimates are much less accurate and more highly variable (\Cref{appendix:lazy_init}). Next, our theory assumes that our training network is over-parameterized and sufficiently wide. We compute empirical confidence intervals for LazyVI for increasing network widths and find that coverage increases as the width increases, but at a computational cost (\Cref{appendix:width}).

\subsection{Binary Classification}
\begin{figure}[ht]
\centerline{\includegraphics[scale=0.22]{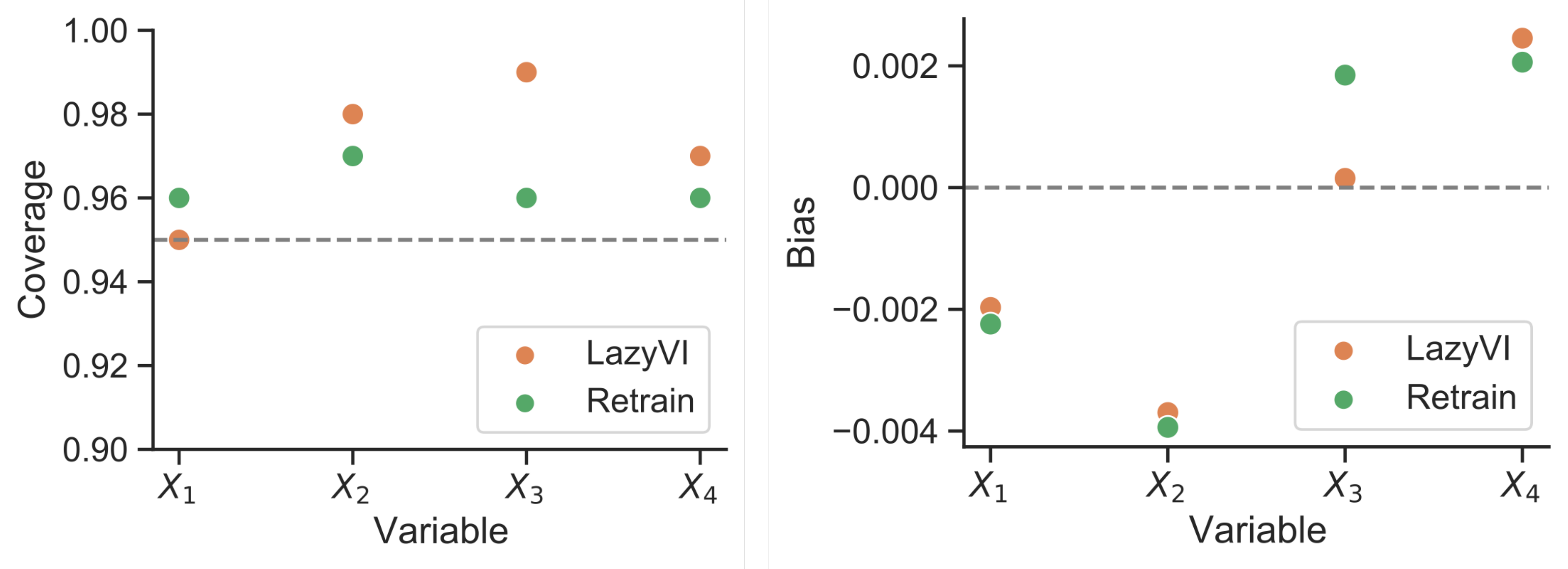}}
\caption{\label{fig:williamson} Left: Average coverage of empirical 95\% confidence intervals from the LazyVI and retrain estimates across 100 simulations. Right: Average empirical bias ($\VI - \hat{\VI}$) of LazyVI and retrain estimates. We see that LazyVI achieves low bias and high coverage in this simulation setting, similar to the Retrain estimates.}
\end{figure}

Because we borrow much of our theoretical framework from \cite{williamson_unified_2020}, we also leverage their simulation framework as a useful point of comparison.
We draw independent samples $X \sim \cN(0,I_{4 \times 4})$ and generate a binary outcome $Y \sim \text{Bernoulli}(\Phi(X\beta))$ where $\beta = (2.5, 3.5, 0, 0)$. Because the outcome is binary, we use accuracy as our predictive skill measure, and the true VI values are given by (0.136, 0.236, 0, 0), respectively. 
We first directly compare the LazyVI and retrain estimators by estimating $\VI$ across 100 simulated datasets of sample size $n=1000$ and computing the empirical 95\% confidence intervals. In Figure \ref{fig:williamson}, we see that the LazyVI and retrain estimates both achieve the desired level of coverage with low bias. 
In this simulation, LazyVI took on average 0.6 seconds (including cross-validating to find the optimal ridge parameter), while retraining took $7.5$ seconds. In this setting, LazyVI is just as accurate as retraining with a more than 10x speed-up.

 \subsection{Nonlinear, High-dimensional Regression} 
\begin{figure}[ht]
\centerline{\includegraphics[scale=0.5]{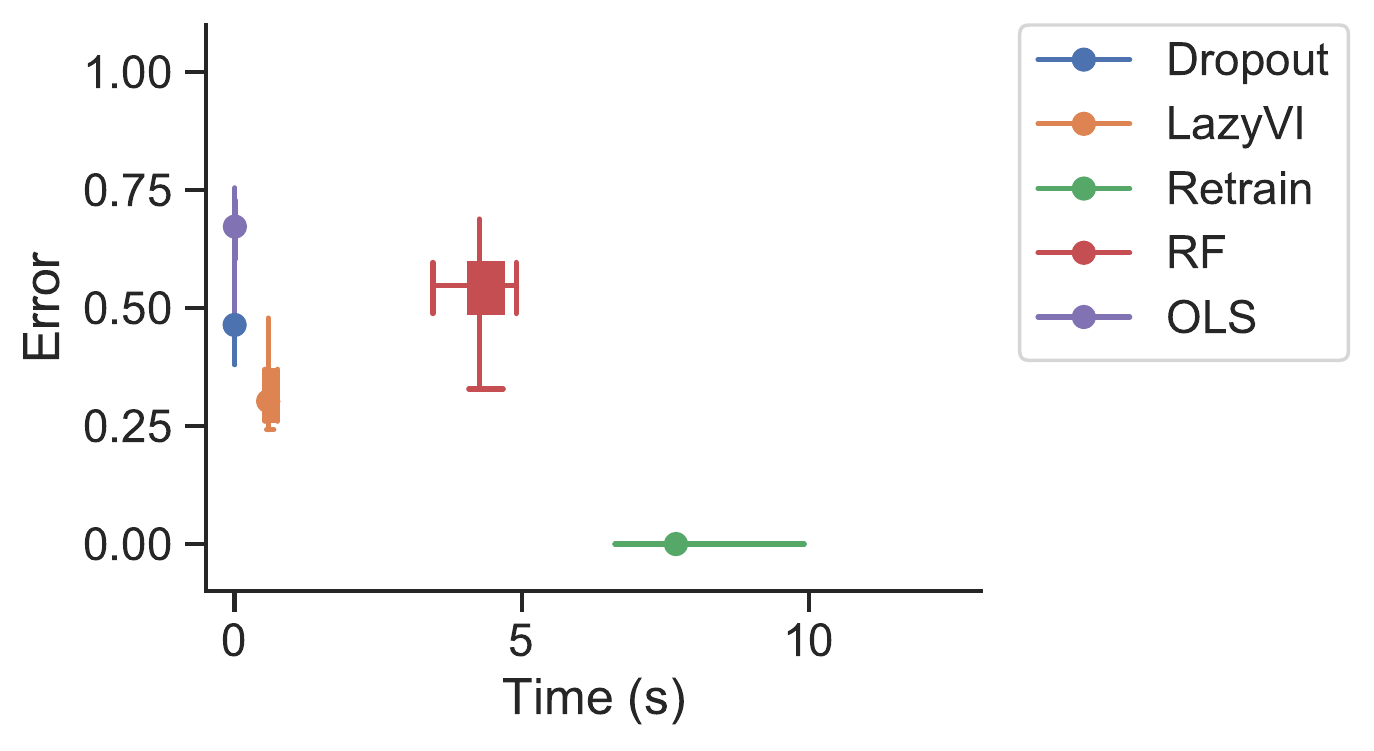}}
\caption{\label{fig:highd}Distribution of computation time vs.\ normalized estimation error relative to retrain for the VI of $X_1$ (($\hat{\VI} - \VIRT)/\VIRT)$ across 10 repetitions. Among the non-Retrain methods, LazyVI is both the fastest and most accurate.}
\end{figure}

The computational burden of retraining is most pronounced in high-dimensional settings, since estimating $\VIRT$ for all variables requires refitting at least $p$ models. For this simulation, we have data $X \sim N(0, \Sigma_{100 \times 100})$, where variables are independent except $\text{Corr}(X_1, X_2) = 0.5$. Letting $\beta = (5, 4, 3, 2, 1, 0, \dots, 0)^{\top} \in \mathbb{R}^{100}$, we construct a weight matrix $W \in \mathbb{R}^{m \times p}$ such that the $W_{:, j} \sim \cN(\beta_j, \sigma^2)$ (i.e. the weights associated with variable $j$ are centered at $\beta_j$). Letting $V \sim \cN(0,1)$, we generate the response $Y_i = V\sigma(W\X_i) + \epsilon_i$ where $\sigma$ is the ReLU function. Because the ``true'' VI values are unknown and difficult to estimate, we present the accuracy of different estimation methods relative to the retraining estimates, which we take as ground truth. We estimate $\VI$ for $X_1$ across 10 simulated datasets ($n=1000$) and benchmark against retraining using both a linear regression (OLS) and random forest (RF).In Figure \ref{fig:highd}, we show the spread of both the computation time and normalized error (relative to retrain) for all methods. We see that LazyVI is the most accurate method and is substantially faster than retraining, which is especially beneficial in this high-dimensional setting.
 
\section{Predicting Seasonal Precipitation}\label{sec:s2s}
Extreme precipitation events have become more and more common in recent years, and are expected to intensify with climate change  \cite{tabari_climate_2020,li_larger_2019}. Early and reliable precipitation forecasting is thus critical for regional water resource management, which increasingly impacts large swaths of the population \cite{aghakouchak_water_2015}. Many studies have shown that the sea surface temperature (SST) over various regions of the ocean, such as the El Niño-Southern Oscillation (ENSO), are predictive of precipitation in the United States \cite{mamalakis_new_2018,dai_influence_2013,lenssen_seasonal_2020}. Understanding which ocean regions are most predictive is challenging, however, due to a short observational record and strong correlations among SSTs \cite{stevens_graph-guided_2021}.

\subsection{Importance of Ocean Climate Indices}
We estimate the importance of different ocean regions for seasonal precipitation forecasting using our lazy training method. The response is the average winter precipitation over the Southwestern US, and as predictors we use 10 ocean climate indices (OCIs), which are defined as the average detrended SST anomalies over different ocean regions \cite{chen_how_2016}. As data, we use simulations from the Community Earth System Model-Large Ensemble project (CESM-LENS; \cite{kay_community_2015,aws_lens}). Details about data  processing can be found in \cref{appx:climate_details}. 

\begin{figure}
\centering
  \includegraphics[scale=.4]{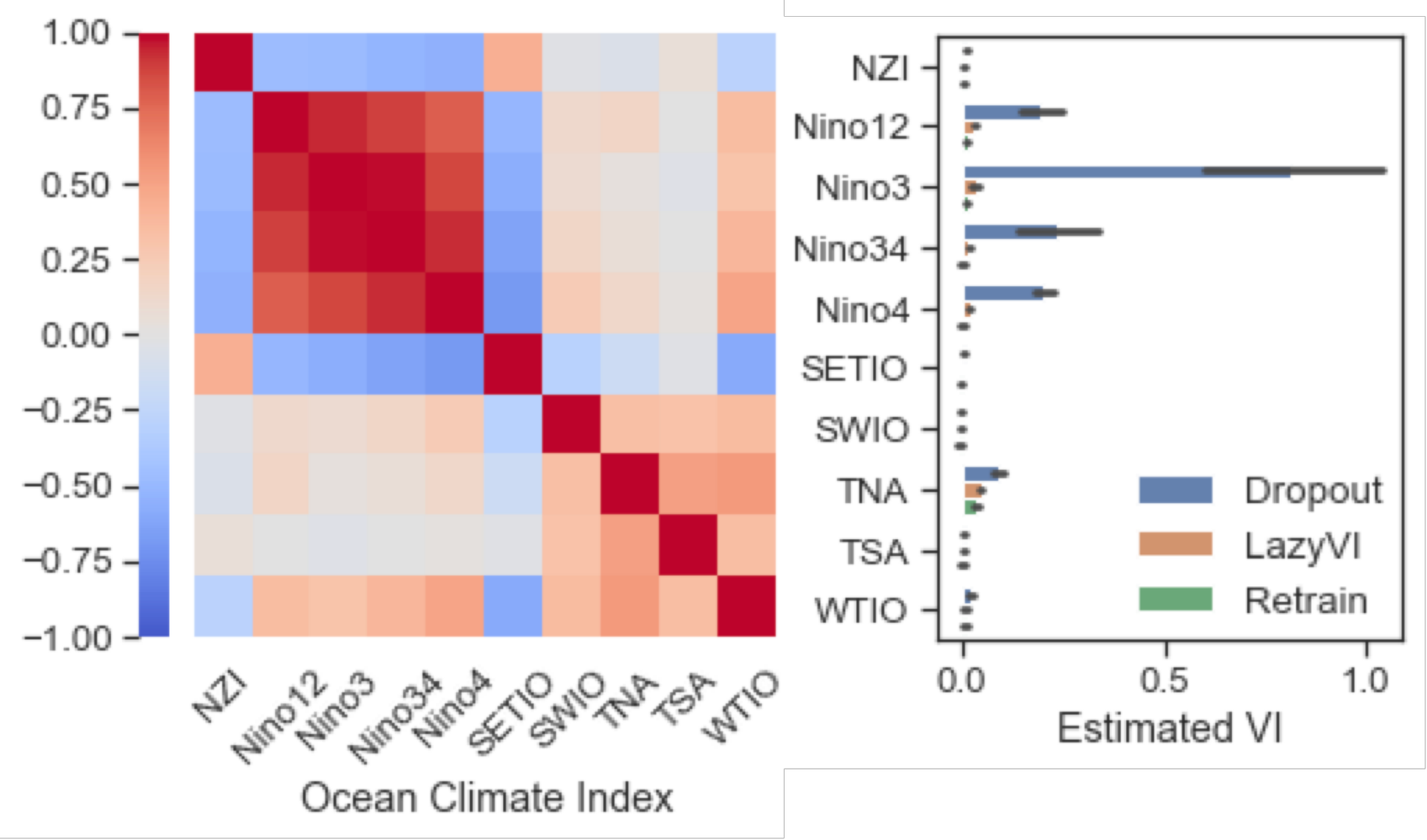}
  \caption{\small \label{oci_panel}Left: sample covariance matrix of the OCIs across the 40 LENS ensemble members; Right: estimated VI for each OCI across 10 different train/test splits. Major discrepancies between Dropout and LazyVI/Retrain occur in regions of high correlation.}
\end{figure}

There are strong correlations among the various OCIs (\Cref{oci_panel})) — in particular, the various Niño indices appear to be nearly collinear. Because of this, we would expect methods like linear regression to inaccurately estimate coefficients and their importance (see appendix for more discussion).

We apply LazyVI to this problem \revise{by first training a two-layer neural network with a hidden width of 50 and then removing each climate index and linearly estimating the correction}. When comparing with the dropout and retraining VI estimates, we see that dropout drastically overestimates VI of Niño 3 and Niño 3.4 relative to retraining, and that LazyVI results in estimates much closer to the retraining estimates. These results are consistent with recent literature indicating that the predictive ability of Niño is often overstated relative to other OCIs \cite{mamalakis_new_2018}, suggesting that LazyVI could potentially help us better 
understand the relative importance of different climate mechanisms.

\begin{figure}[!ht]
\centerline{\includegraphics[scale=0.25]{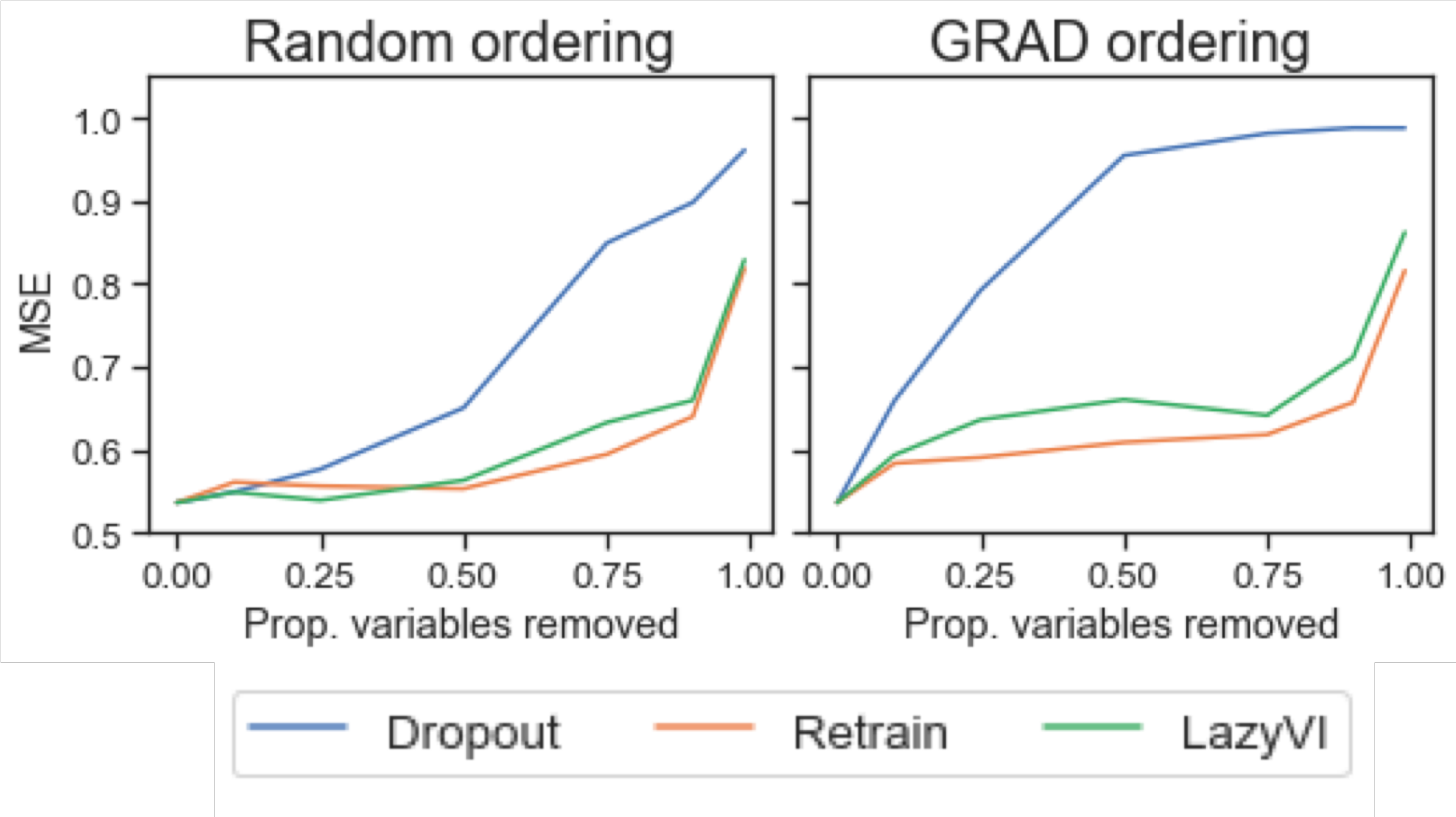}}
\caption{\label{fig:cli_roar} Average MSE across 5 runs after removing increasing proportions of variables with the specified importance orderings (see \cref{fig:saliency} for more detail). On average, LazyVI took 1.8s and Retrain took 3.2s.}
\end{figure}

\revise{
\subsection{High-dimensional Seasonal Forecasting}\label{sec:hdcli}
Aggregating climate regions into OCIs is standard in the climate literature and a critical tool for understanding climate dynamics. However, while more difficult to interpret and estimate, disaggregating OCIs and investigating individual SST locations offers important insights into the rapidly changing climate system \cite{stevens_graph-guided_2021}. Neural networks have increasingly been used to make these types of high-dimensional forecasts, and with that comes an increased interest in explainability \cite{mamalakis_ebert-uphoff_barnes_2022}. However, standard gradient-based attribution/saliency methods used to interpret NNs, while powerful for particular networks, are often subjective and  difficult to interpret \cite{mamalakis-xai-22}. 

The ROAR (RemOve and Retrain) framework introduced by \cite{hooker_benchmark_2019} offers a helpful way to evaluate such importance measures. This work provides a retraining-based benchmark for evaluating NN attribution/saliency methods by removing variables in order of estimated importance and measuring the drop in predictive power. This work finds that many common attribution methods are no more informative than a random baseline, and aruges that retraining the network after dropping out variables is key in understanding this behavior.

Using all summer SSTs across the Pacific basin on a $10^{\circ} \times 10^{\circ}$ grid, for a total of 220 predictors, we show that LazyVI can achieve similar results to retraining in the ROAR framework at a computational speed-up. We train fully-connected three-layer neural network of widths (100, 50) on all variables, and from this trained network, we estimate feature importance using the baseline Gradients importance method (GRAD, \cite{simonyan}). We then remove $t=(.1, .25, .5, .75, .9, .99)$ proportion of variables by removing them in order of GRAD importance (in addition to a random ordering as a baseline, see \cref{fig:saliency} in \cref{appx:climate_details} for a visualization of this procedure). We estimate model performance on these modified datasets using the Dropout, Retraining (ROAR) and LazyVI approaches and find that LazyVI closely approximates ROAR in nearly half the time (Figure \ref{fig:cli_roar}) and the ordering of variables removed does not matter much; Dropout, on the other hand, vastly overestimates the degradation of the model performance, even with a small number of variables removed.
}

\section{Discussion and Extended Applications}

Assessing variable importance in machine learning is a vital task as learning-based tools are increasingly integrated into societally-impactful systems, including autonomous vehicles, financial and healthcare decision-making, and social and criminal justice.
In this work, we propose a method, LazyVI, for efficiently estimating variable importance based on a linearization of a fully trained neural network. We prove that our method provides an accurate estimate of VI and
can achieve the same rate of accuracy as a computationally expensive retraining method nearly as quickly as the inaccurate dropout method. We further show how to construct confidence intervals around these estimates. 

The theory developed in this paper provides an important step toward making interpretability in neural networks more computationally efficient, and we suspect this theoretical framework can extend to other settings, which we discuss here.

\subsection{Early Stopping and Regularization}
\begin{figure}
\centering
  \includegraphics[scale=.2]{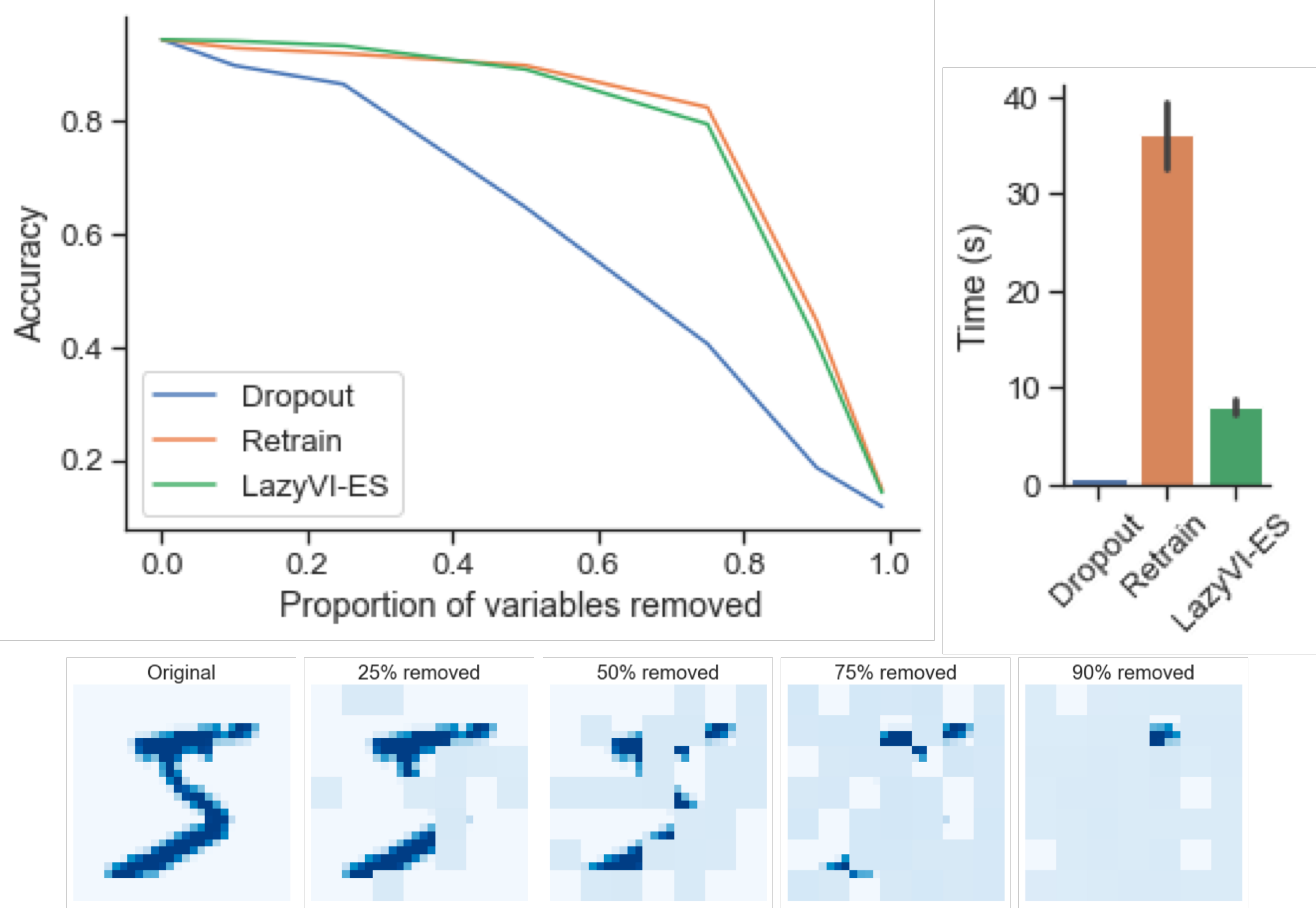}
  \caption{\small \label{mnist_roar} (Left) Average accuracy of estimator across 5 repetitions. (Right) Average computation time for a single run of each method. (Bottom) Example MNIST image with different proportions of variables removed. Early stopping is able to nearly identically mimic a full model retrain at a fraction of the time.}
\end{figure}

A potential alternative to our proposed LazyVI method is to first train a full model (as we do) and then train the reduced model using a gradient-based method initialized with the full model parameters and stopped early. Empirical evidence suggests that this approach would have similar speed and accuracy to our LazyVI approach due to the implicit regularization associated with early stopping. This approach has the potential to extend LazyVI to far more complicated architectures than the standard feedforward networks we have experimented with thus far. As a proof of concept, we train a convolutional neural network on the MNIST benchark dataset, and then follow the ROAR procedure with random ordering. 
We see in Figure \ref{mnist_roar} that Dropout results in a consistent decline in predictive performance when variables are removed, while the accuracy remains relatively high when the network is retrained up until around 75\% of variables are removed - remarkable, given how relatively uninformative the image appears (Figure \ref{mnist_roar}, bottom). Importantly, we see that taking a single step from a model initialized at the full model parameters (LazyVI-ES) results in nearly identical performance to the full retraining at a $5\times$ speed-up.

While these results are promising, we currently lack theoretical guarantees for early stopping in this setting. It is possible that
our theoretical results could lead to new insights into early stopping 
for assessing VI due to the  intimate connection between kernel ridge regression and early stopping algorithms \cite{raskutti2013early}.
In fact, if the eigenvalues of the NTK matrix at the full model initialization decay in a sufficiently fast rate,  early stopping of the 
reduced model training should give an as good estimate of the reduced model. However, analyzing early stopping in this setting requires characterizing the spectrum  of the NTK with \emph{the full model initialization},
whereas most spectral properties of the NTK have been developed under the assumption of a \textit{random initialization}
\cite{nguyen2021tight, montanari2020interpolation}.
Better understanding the NTK spectrum after full-model initialization in the future could provide new insights into fast algorithms for VI estimation.

\subsection{Shapley Values}
When features are correlated, the quantity VI defined in  \eqref{eq:VIdef} tends to zero. 
Recent work proposes using Shapley values to measure variable importance, arguing that their handling of correlated variables, which assigns similar positive weights to correlated important variables, is desirable \cite{owen_shapley_2016,williamson_efficient_2020} in some settings.
These papers also note that Shapley values are prohibitively expensive to compute, as they require fitting a new model for each of the $2^p$ possible subsets of variables. 
However, we note that computing the Shapley values requires many calculations of the quantity in \eqref{eq:VIdef}; an important avenue is investigating the use of our LazyVI framework to accelerate the computation of Shapley values. 
\revise{ We perform a preliminary experiment on calculating Shapley values using our LazyVI framework, and compare it with the retraining method used in \cite{williamson_efficient_2020}. We estimate the Shapley values for a sparse high-dimensional data generated by a logistic model, and perform the retraining/lazyVI method on a two-layer neural network. When using LazyVI training, the computation is roughly $5$ times faster and the estimated Shapley values are close to retraining. Moreover, when the sample size is relatively small with respective to the dimension, we observe Lazy training has a smaller variance of estimated Shapley values on non-important variables than retraining method, due to the regularization proposed in our method.
See 
\cref{appdx: shapley} for our initial exploration into this line of work.}

\section*{Acknowledgements}
This work was supported by AFOSR FA9550-18-1-0166, DOE DE-AC02-06CH113575, NSF OAC-1934637, NSF DMS-1930049 and NSF DMS-2023109.

\bibliographystyle{alpha}
\bibliography{vi_refs}

\newcommand{\etalchar}[1]{$^{#1}$}
\begin{thebibliography}{SWM{\etalchar{+}}21}

\bibitem[AFH{\etalchar{+}}15]{aghakouchak_water_2015}
Amir AghaKouchak, David Feldman, Martin Hoerling, Travis Huxman, and Jay Lund.
\newblock Water and climate: {Recognize} anthropogenic drought.
\newblock {\em Nature}, 524(7566), August 2015.

\bibitem[Bar94]{barron1994approximation}
Andrew~R Barron.
\newblock Approximation and estimation bounds for artificial neural networks.
\newblock {\em Machine learning}, 14(1):115--133, 1994.

\bibitem[BBM{\etalchar{+}}15]{bach2015pixel}
Sebastian Bach, Alexander Binder, Gr{\'e}goire Montavon, Frederick Klauschen,
  Klaus-Robert M{\"u}ller, and Wojciech Samek.
\newblock On pixel-wise explanations for non-linear classifier decisions by
  layer-wise relevance propagation.
\newblock {\em PloS one}, 10(7):e0130140, 2015.

\bibitem[BC18]{barber2018knockoff}
Rina~Foygel Barber and Emmanuel~J. Candes.
\newblock A knockoff filter for high-dimensional selective inference, 2018.

\bibitem[BM02]{bartlett2002rademacher}
Peter~L Bartlett and Shahar Mendelson.
\newblock Rademacher and gaussian complexities: Risk bounds and structural
  results.
\newblock {\em Journal of Machine Learning Research}, 3(Nov):463--482, 2002.

\bibitem[CFJL17]{candes_panning_2017}
Emmanuel Candes, Yingying Fan, Lucas Janson, and Jinchi Lv.
\newblock Panning for {Gold}: {Model}-{X} {Knockoffs} for {High}-dimensional
  {Controlled} {Variable} {Selection}.
\newblock {\em arXiv:1610.02351 [math, stat]}, December 2017.
\newblock arXiv: 1610.02351.

\bibitem[CMA{\etalchar{+}}16]{chen_how_2016}
Yang Chen, Douglas~C. Morton, Niels Andela, Louis Giglio, and James~T.
  Randerson.
\newblock How much global burned area can be forecast on seasonal time scales
  using sea surface temperatures?
\newblock {\em Environmental Research Letters}, 11(4):045001, March 2016.

\bibitem[COB20]{chizat_lazy_2020}
Lenaic Chizat, Edouard Oyallon, and Francis Bach.
\newblock On {Lazy} {Training} in {Differentiable} {Programming}.
\newblock {\em arXiv:1812.07956}, January 2020.

\bibitem[CRG17]{chang2017dropout}
Chun-Hao Chang, Ladislav Rampasek, and Anna Goldenberg.
\newblock Dropout feature ranking for deep learning models.
\newblock {\em arXiv preprint arXiv:1712.08645}, 2017.

\bibitem[Dai13]{dai_influence_2013}
Aiguo Dai.
\newblock The influence of the inter-decadal {Pacific} oscillation on {US}
  precipitation during 1923–2010.
\newblock {\em Climate Dynamics}, 41(3), August 2013.

\bibitem[dBS{\etalchar{+}}19]{aws_lens}
J.~{de La Beaujardi{\`e}re}, A.~{Banihirwe}, C.~F.~G. {Shih}, K.~{Paul}, and
  J.~{Hamman}.
\newblock Ncar cesm lens cloud-optimized subset.
\newblock {\em UCAR/NCAR Computational and Informations Systems Lab}, 2019.

\bibitem[DS95]{doksum1995nonparametric}
Kjell Doksum and Alexander Samarov.
\newblock Nonparametric estimation of global functionals and a measure of the
  explanatory power of covariates in regression.
\newblock {\em The Annals of Statistics}, pages 1443--1473, 1995.

\bibitem[GMR{\etalchar{+}}18]{Guidotti_surveybbb}
Riccardo Guidotti, Anna Monreale, Salvatore Ruggieri, Franco Turini, Fosca
  Giannotti, and Dino Pedreschi.
\newblock A survey of methods for explaining black box models.
\newblock {\em ACM Comput. Surv.}, 51(5), August 2018.

\bibitem[H{\etalchar{+}}19]{hooker_benchmark_2019}
Sara Hooker et~al.
\newblock A benchmark for interpretability methods in deep neural networks.
\newblock In {\em NeurIPS}, 2019.

\bibitem[HKZ12]{hsu2012tail}
Daniel Hsu, Sham Kakade, and Tong Zhang.
\newblock A tail inequality for quadratic forms of subgaussian random vectors.
\newblock {\em Electronic Communications in Probability}, 17:1--6, 2012.

\bibitem[HLY19]{hu2019simple}
Wei Hu, Zhiyuan Li, and Dingli Yu.
\newblock Simple and effective regularization methods for training on noisily
  labeled data with generalization guarantee.
\newblock {\em arXiv preprint arXiv:1905.11368}, 2019.

\bibitem[JGH20]{jacot2020neural}
Arthur Jacot, Franck Gabriel, and Clément Hongler.
\newblock Neural tangent kernel: Convergence and generalization in neural
  networks, 2020.

\bibitem[KDP{\etalchar{+}}15]{kay_community_2015}
J.~E. Kay, C.~Deser, A.~Phillips, A.~Mai, C.~Hannay, G.~Strand, J.~M.
  Arblaster, S.~C. Bates, G.~Danabasoglu, J.~Edwards, M.~Holland, P.~Kushner,
  J.-F. Lamarque, D.~Lawrence, K.~Lindsay, A.~Middleton, E.~Munoz, R.~Neale,
  K.~Oleson, L.~Polvani, and M.~Vertenstein.
\newblock The {Community} {Earth} {System} {Model} ({CESM}) {Large} {Ensemble}
  {Project}: {A} {Community} {Resource} for {Studying} {Climate} {Change} in
  the {Presence} of {Internal} {Climate} {Variability}.
\newblock {\em Bulletin of the American Meteorological Society},
  96(8):1333--1349, August 2015.

\bibitem[LGM20]{lenssen_seasonal_2020}
Nathan J.~L. Lenssen, Lisa Goddard, and Simon Mason.
\newblock Seasonal {Forecast} {Skill} of {ENSO} {Teleconnection} {Maps}.
\newblock {\em Weather and Forecasting}, 35(6):2387--2406, December 2020.
\newblock Publisher: American Meteorological Society Section: Weather and
  Forecasting.

\bibitem[LGR{\etalchar{+}}18]{jinglei2018}
Jing Lei, Max G’Sell, Alessandro Rinaldo, Ryan~J. Tibshirani, and Larry
  Wasserman.
\newblock Distribution-free predictive inference for regression.
\newblock {\em Journal of the American Statistical Association},
  113(523):1094--1111, 2018.

\bibitem[LXS{\etalchar{+}}20]{2020Lee}
Jaehoon Lee, Lechao Xiao, Samuel~S Schoenholz, Yasaman Bahri, Roman Novak,
  Jascha Sohl-Dickstein, and Jeffrey Pennington.
\newblock Wide neural networks of any depth evolve as linear models under
  gradient descent.
\newblock {\em Journal of Statistical Mechanics: Theory and Experiment},
  2020(12):124002, Dec 2020.

\bibitem[LZZ{\etalchar{+}}19]{li_larger_2019}
Chao Li, Francis Zwiers, Xuebin Zhang, Gang Chen, Jian Lu, Guilong Li, Jesse
  Norris, Yaheng Tan, Ying Sun, and Min Liu.
\newblock Larger {Increases} in {More} {Extreme} {Local} {Precipitation}
  {Events} as {Climate} {Warms}.
\newblock {\em Geophysical Research Letters}, 46(12):6885--6891, 2019.

\bibitem[MBEU22]{mamalakis-xai-22}
Antonios Mamalakis, Elizabeth~A. Barnes, and Imme Ebert-Uphoff.
\newblock Investigating the fidelity of explainable artificial intelligence
  methods for applications of convolutional neural networks in geoscience,
  2022.

\bibitem[MEUB22]{mamalakis_ebert-uphoff_barnes_2022}
Antonios Mamalakis, Imme Ebert-Uphoff, and Elizabeth~A. Barnes.
\newblock Neural network attribution methods for problems in geoscience: A
  novel synthetic benchmark dataset.
\newblock {\em Environmental Data Science}, 1:e8, 2022.

\bibitem[MRT18]{mohri2018foundations}
Mehryar Mohri, Afshin Rostamizadeh, and Ameet Talwalkar.
\newblock {\em Foundations of machine learning}.
\newblock 2018.

\bibitem[MYR{\etalchar{+}}18]{mamalakis_new_2018}
Antonios Mamalakis, Jin-Yi Yu, James~T. Randerson, Amir AghaKouchak, and Efi
  Foufoula-Georgiou.
\newblock A new interhemispheric teleconnection increases predictability of
  winter precipitation in southwestern {US}.
\newblock {\em Nature Communications}, 9(1):2332, June 2018.

\bibitem[MZ20]{montanari2020interpolation}
Andrea Montanari and Yiqiao Zhong.
\newblock The interpolation phase transition in neural networks: Memorization
  and generalization under lazy training.
\newblock {\em arXiv preprint arXiv:2007.12826}, 2020.

\bibitem[NMM21]{nguyen2021tight}
Quynh Nguyen, Marco Mondelli, and Guido~F Montufar.
\newblock Tight bounds on the smallest eigenvalue of the neural tangent kernel
  for deep relu networks.
\newblock In {\em International Conference on Machine Learning}, pages
  8119--8129. PMLR, 2021.

\bibitem[OP16]{owen_shapley_2016}
Art~B. Owen and Clémentine Prieur.
\newblock On {Shapley} value for measuring importance of dependent inputs,
  October 2016.

\bibitem[RR19]{Rudin2019Why}
Cynthia Rudin and Joanna Radin.
\newblock Why are we using black box models in ai when we don’t need to? a
  lesson from an explainable ai competition.
\newblock {\em Harvard Data Science Review}, 1(2), 11 2019.
\newblock https://hdsr.mitpress.mit.edu/pub/f9kuryi8.

\bibitem[RWY14]{raskutti2013early}
Garvesh Raskutti, Martin~J. Wainwright, and Bin Yu.
\newblock Early stopping and non-parametric regression: An optimal
  data-dependent stopping rule.
\newblock {\em Journal of Machine Learning Research}, 2014.

\bibitem[SGK19]{shrikumar_learning_2019}
Avanti Shrikumar, Peyton Greenside, and Anshul Kundaje.
\newblock Learning {Important} {Features} {Through} {Propagating} {Activation}
  {Differences}.
\newblock {\em arXiv:1704.02685 [cs]}, October 2019.
\newblock arXiv: 1704.02685.

\bibitem[STK{\etalchar{+}}17]{smilkov2017smoothgrad}
Daniel Smilkov, Nikhil Thorat, Been Kim, Fernanda Vi{\'e}gas, and Martin
  Wattenberg.
\newblock Smoothgrad: removing noise by adding noise.
\newblock {\em arXiv preprint arXiv:1706.03825}, 2017.

\bibitem[STY17]{sundararajan2017axiomatic}
Mukund Sundararajan, Ankur Taly, and Qiqi Yan.
\newblock Axiomatic attribution for deep networks.
\newblock In {\em International Conference on Machine Learning}, pages
  3319--3328. PMLR, 2017.

\bibitem[SvdLP14]{sapp2014targeted}
Stephanie Sapp, Mark~J van~der Laan, and Kimberly Page.
\newblock Targeted estimation of binary variable importance measures with
  interval-censored outcomes.
\newblock {\em The international journal of biostatistics}, 10(1):77--97, 2014.

\bibitem[SVZ13]{simonyan}
Karen Simonyan, Andrea Vedaldi, and Andrew Zisserman.
\newblock Deep inside convolutional networks: Visualising image classification
  models and saliency maps, 2013.

\bibitem[SWM{\etalchar{+}}21]{stevens_graph-guided_2021}
Abby Stevens, Rebecca Willett, Antonios Mamalakis, Efi Foufoula-Georgiou,
  Alejandro Tejedor, James~T. Randerson, Padhraic Smyth, and Stephen Wright.
\newblock Graph-{Guided} {Regularized} {Regression} of {Pacific} {Ocean}
  {Climate} {Variables} to {Increase} {Predictive} {Skill} of {Southwestern}
  {U}.{S}. {Winter} {Precipitation}.
\newblock {\em Journal of Climate}, 34(2):737--754, January 2021.

\bibitem[Tab20]{tabari_climate_2020}
Hossein Tabari.
\newblock Climate change impact on flood and extreme precipitation increases
  with water availability.
\newblock {\em Scientific Reports}, 10(1), August 2020.

\bibitem[WF20]{williamson_efficient_2020}
Brian~D. Williamson and Jean Feng.
\newblock Efficient nonparametric statistical inference on population feature
  importance using {Shapley} values.
\newblock {\em arXiv:2006.09481 [stat]}, June 2020.
\newblock arXiv: 2006.09481.

\bibitem[WGSC21]{williamson_unified_2020}
Brian~D. Williamson, Peter~B. Gilbert, Noah~R. Simon, and Marco Carone.
\newblock A general framework for inference on algorithm-agnostic variable
  importance.
\newblock {\em arXiv:2004.03683}, 2021.

\bibitem[ZJ21]{zhang_floodgate_2021}
Lu~Zhang and Lucas Janson.
\newblock Floodgate: inference for model-free variable importance.
\newblock {\em arXiv:2007.01283 [stat]}, April 2021.
\newblock arXiv: 2007.01283.

\end{thebibliography}

\newpage

\begin{appendices}

\section{Supporting Lemmas and Proofs}

\subsection{Supporting Lemma}

\begin{assumption}
\begin{itemize}
    \item[(A1)] There exists some constant $C>0$ such that, for each sequence $f_1, f_2, \dots \in \mathcal{F}$ such that $\|f_i - f_0\|_{\mathcal{F}} \rightarrow 0$,
    $| V(f_j, P_0) - V(f_0, P_0) | \leq C \| f_j - f_0\|_{\mathcal{F}}^2$ for each $j$ large enough;
    \item[(A2)] There exists some constant $\delta>0$ such that for each sequence $\epsilon_1, \epsilon_2, \dots\in \bbR$ and $h, h_1, h_2,\dots \in \bbR$ satisfying that $\epsilon_j\rightarrow 0$ and $\| h_j - h\|_{\infty} \rightarrow 0$, it holds that 
    \begin{equation*}
        \sup_{f\in \mathcal{F}: \|f-f_0\|_{\mathcal{F}} <\delta} | \frac{V(f, P_0+\epsilon_j h_j) - V(f, P_0)}{\epsilon_j} - \dot{V}(f, P_0; h_j)| \rightarrow 0;
    \end{equation*}
    \item[(B2)] $\int [g_n(z)]^2 dP_0(z) = o_P(1) $;
\end{itemize}

\end{assumption}

\begin{lemma}(\cite{williamson_unified_2020})
    Suppose (A1-A2, B2) regularity conditions hold. Denote $f_n(X)$ and $f_{n,-j}(X_{-j})$ as the estimate for $f_{0}$ and $f_{0,-j}$,  
    Then for a predictive skill measure $V(f,P)$ satisfying conditions (A1)-(A2), (B2) in Appendix, as long as the estimators satisfy the following condition:
    \begin{equation}
        \| f_n - f_0\|_{\mathcal{F}} = O_p(n^{-\frac{1}{4}}), ~~ 
        \| f_{n,-j} - f_{0,-j}\|_{\mathcal{F}} = O_p(n^{-\frac{1}{4}}), 
    \end{equation}
    for all $j\in[p]$, then
    we have
    \begin{equation}
        \begin{split}
            v_n- v_0 =& \frac{1}{n}\sum_{i=1}^n \dot{V} (f_0, P_0; \delta_{Z_i} - P_0) + O_p(\frac{1}{\sqrt{n}}),\\
            v_{n,-j} - v_{0,-j} = & \frac{1}{n}\sum_{i=1}^n \dot{V} (f_{0,-j}, P_{0,-j}; \delta_{Z_i} - P_{0,-j}) 
            + O_p(\frac{1}{\sqrt{n}}), 
        \end{split}
    \end{equation}
    where  $v_n = V(f_n,P_n)$ and $v_{n,-j} =  V(f_{n,-j}, P_{n,-j}), ~\forall j\in [p]$.
    \label{lemma: williamson}
\end{lemma}

\subsection{Missing Proofs}

\subsubsection{Proof of \Cref{lemma: KRR}}

\textit{Here we present the detailed proof of \Cref{lemma: KRR}, which gives the empirical estimation error bounds for the NTK kernel ridge regression estimation. The proof follows the proof framework provided in \cite{hu2019simple}}.

First of all, according to kernel ridge regression, 
denote 
\begin{itemize}
    \item $\Y = (Y_1, \dots, Y_n)^\top$;
    \item $\bfhj = (h_{\theta_f}(\X_1^{(j)}),\dots, h_{\theta_f}(\X_n^{(j)}) )^\top$;
    \item $\bffj = (f_{0,-j}(\X_1^{(j)}), \dots, f_{0,-j}(\X_n^{(j)}))^\top$.
\end{itemize}

we have 
\begin{equation}
    (\tilde{h}_{\theta_f + \Delta \theta_j} (\X_1^{(j)}), \dots, \tilde{h}_{\theta_f + \Delta \theta_j} (\X_n^{(j)} ))^\top = \bbK^{(j)} (\bbK^{(j)} + \lambda I_n)^{-1} (\Y - \bfhj) + \bfhj.
\end{equation}

Recall that $\epsilon^{(j)} = Y - \bbE(Y|X_{-j}) = Y - f_{0,-j}(\X^{(j)})$, we define its observed samples as
$$\bmepsj = \left(Y_1 - f_{0,-j}(\X_1^{(j)}),\dots,Y_n - f_{0,-j}(\X_n^{(j)})\right)^\top = \Y - \bffj;$$
Recall the definition of $\e^{(j)}$, we have 
$$\e^{(j)} = \bffj - \bfhj.$$

Hence we have 
\begin{equation}
    \begin{split}
        &\sqrt{n}\| \linearh (X^{(j)}) - f_{0,-j} (X^{(j)})\|_n\\
        = & \sqrt{  \sum_{i = 1}^n 
        \left[\linearh(\X_i^{(j)}) - f_{0,-j}(\X_i^{(j)}) \right]^2
        }\\
        = & \| \bbK^{(j)} (\bbK^{(j)} + \lambda I_n)^{-1} ( \Y - \bfhj ) + \bfhj - \bffj\| \\
        = & \| \bbK^{(j)} (\bbK^{(j)} + \lambda I_n)^{-1} ( \bffj + \bmepsj - \bfhj ) + \bfhj - \bffj\| \\
        = &\| \bbK^{(j)} (\bbK^{(j)} + \lambda I_n)^{-1} ( \e^{(j)} + \bmepsj )  - \e^{(j)}\|\\
        = &\| \bbK^{(j)} (\bbK^{(j)} + \lambda I_n)^{-1}   \bmepsj  -\lambda (\bbK^{(j)} + \lambda I_n)^{-1} \e^{(j)}\|\\
        \leq & \|  \bbK^{(j)} (\bbK^{(j)} + \lambda I_n)^{-1}   \bmepsj\| + \|\lambda (\bbK^{(j)} + \lambda I_n)^{-1} \e^{(j)}\|.
    \end{split}
    \label{eq: bound linearh}
\end{equation}

According to \cref{asspt: tail bound} and \cite{hsu2012tail}, we have
\begin{equation}
    P\left(\|A\bmepsj\|^2/\sigma^2 >  \tr(\Sigma) + 2 \sqrt{\tr(\Sigma^2) t} + 2 \|\Sigma\| t ~|~ \X^{(j)} \right) \leq e^{-t},
\end{equation}
where $A = \bbK^{(j)} (\bbK^{(j)} + \lambda I_n)^{-1}$, and $\Sigma = A^\top A$. Hence we have with probability at least $1-\delta$ for any $\delta >0$, we have 
\begin{equation}
    \|  \bbK^{(j)} (\bbK^{(j)} + \lambda I_n)^{-1}   \bmepsj\| \leq \sigma \sqrt{\tr(\Sigma) + 2 \sqrt{\tr(\Sigma^2)\log(\frac{1}{\delta})} +2\|\Sigma\|\log(\frac{1}{\delta}) }.
\end{equation}
Let $\lambda_1, \dots, \lambda_n>0$ be the eigenvalues of $\bbK^{(j)}$, we then have
\begin{equation}
    \begin{split}
        &\tr[\Sigma] = \tr[A^\top A] = \sum_{i=1}^n \frac{\lambda_i^2}{(\lambda_i + \lambda)^2} \leq \sum_{i=1}^n \frac{\lambda_i^2}{4\lambda_i \cdot \lambda} = \frac{\tr[\bbK^{(j)}]}{4\lambda},\\
        &\tr[\Sigma^2] =\tr[A^\top A^2 A^\top]= \sum_{i=1}^n \frac{\lambda_i^4}{(\lambda_i + \lambda)^4} \leq \sum_{i=1}^n \frac{\lambda_i^4}{4^4 \lambda (\frac{\lambda_i}{3})^3} = \frac{3^3\tr[\bbK^{(j)}]}{4^4\lambda} \leq \frac{\tr[\bbK^{(j)}]}{4\lambda},\\
        & \|\Sigma\| = \| \bbKj (\bbKj + \lambda I_n)^{-2} \bbKj\| \leq 1.
    \end{split}
\end{equation}

Hence we have with probability at least $1-\delta$,
\begin{equation}
    \begin{split}
         &\|  \bbK^{(j)} (\bbK^{(j)} + \lambda I_n)^{-1}   \bmepsj\| \\
         \leq & \sigma\sqrt{\frac{\tr[\bbKj]}{4\lambda} + 2\sqrt{\frac{\tr[\bbKj]}{4\lambda}} + 2\log(\frac{1}{\delta}) }\\
         \leq & \sigma\sqrt{\frac{\tr[\bbKj]}{4\lambda} + 2\sqrt{\frac{\tr[\bbKj]}{2\lambda}} + 2\log(\frac{1}{\delta}) }\\
         =& \sigma \sqrt{\frac{\tr[\bbKj]}{4\lambda}} + \sigma \sqrt{2\log(\frac{1}{\delta})}.
    \end{split}
\end{equation}

\revise{By the fact that}
\begin{equation}
    \|\lambda (\bbK^{(j)} + \lambda I_n)^{-1} \e^{(j)}\| 
    = \lambda \sqrt{(\e^{(j)})^\top (\bbK^{(j)} + \lambda I_n)^{-2} \e^{(j)}},
\end{equation}
\revise
{we have}
\begin{equation}
    \| \linearh (X^{(j)}) - f_{0,-j} (X^{(j)})\|_n \leq \frac{{\lambda}}{\sqrt{n}} \sqrt{(\e^{(j)})^\top [\bbK^{(j)} + \lambda I_n]^{-2} \e^{(j)}} + \sigma \sqrt{\frac{\tr[\bbKj]}{4n \lambda}} + \sigma \sqrt{\frac{2}{n}\log(\frac{1}{\delta})}.
\end{equation}

\hfill$\square$

\subsubsection{\Cref{lemma: Hnorm_bound} and Its Proof}
Define the Hilbert norm for a function $f(x) = \alpha^\top K(x, \X^{(j)}), ~\forall \alpha\in \bbR^n$ in the NTK kernel space is: $\| f\|_{\mathcal{H}} = \sqrt{\alpha^\top \bbKj \alpha}$.
The following lemma is to bound the Hilbert norm for $\linearh - h_{\theta_f}$ so that we could bound the complexity of the function class it lies in.

\begin{lemma}
With probability at least $1 - \delta$, for any $j\in [p]$ we have
    \begin{equation}
        \| \linearh(x) - h_{\theta_f}(x) \|_{\cH} \leq \sqrt{(\ej )^\top (\bbKj + \lambda I_n)^{-1} \ej  } + \frac{\sigma}{\sqrt{\lambda}} \left(\sqrt{n} + \sqrt{2\log(\frac{1}{\delta})}
    \right).
    \end{equation}
    \label{lemma: Hnorm_bound}
\end{lemma}

\textit{[Proof.]}
Recall that $\linearh(x) = \kerf (x, \X^{(j)} ) (\bbK^{(j)} +\lambda I_n )^{-1} (\Y - \bfhj) + h_{\theta_f}(x) $.
Based on the fact that $\Y - \bfhj = \ej + \bmepsj$, we have
\begin{equation}
    \begin{split}
        &\| \linearh(x) - h_{\theta_f}(x) \|_{\cH}\\
        =& \| (\Y - \bfhj)^\top (\bbK^{(j)} +\lambda I_n )^{-1}  \kerf ( \X^{(j)}, x )\|_{\cH}\\
        =& \sqrt{(\ej + \bmepsj)^\top (\bbKj + \lambda I_n)^{-1}\bbKj (\bbKj + \lambda I_n)^{-1} (\ej + \bmepsj) }\\
        \leq & \sqrt{(\ej + \bmepsj)^\top (\bbKj + \lambda I_n)^{-1} (\ej + \bmepsj) }\\
        \leq & \sqrt{(\ej )^\top (\bbKj + \lambda I_n)^{-1} \ej  } + \sqrt{(\bmepsj)^\top (\bbKj + \lambda I_n)^{-1}  \bmepsj }\\
        \leq& \sqrt{(\ej )^\top (\bbKj + \lambda I_n)^{-1} \ej  } + \sqrt{\frac{(\bmepsj)^\top  \bmepsj} {\lambda}}.
    \end{split}
    \label{eq: Hnorm_sep}
\end{equation}

Using the concentration inequality in \cite{hsu2012tail} again, we have with probability at least $1-\delta$,
we have
\begin{equation}
    \sqrt{(\bmepsj)^\top \bmepsj} \leq \sigma \sqrt{n + 2 \sqrt{n\log(\frac{1}{\delta})} + 2\log(\frac{1}{\delta})}
    \leq \sigma \left(\sqrt{n} + \sqrt{2\log(\frac{1}{\delta})}
    \right).
    \label{eq: bmepsj}
\end{equation}

Hence we prove \Cref{lemma: Hnorm_bound} by combining \cref{eq: Hnorm_sep} and \cref{eq: bmepsj}.

\subsubsection{Generalization Error Bound and Its Proof}
\label{appendix: generalization_error}
\textit{In the following, we will bound the generalization error based on the above empirical error bound.}

\begin{lemma}\label{lemma: ridge generalization bound}
For any $j\in [p]$, let $\|\cdot\|$ be the $L_2(P_{0})$ norm defined as $\|f\| = \sqrt{\int |f(x^{(j)})| dP_{0}(x)}$, then we have with probability at least $1-\delta$ for any $\delta>0$, 
    \begin{equation}
        \begin{split}
            \|\linearh - f_{0,-j}\| \leq &\left\{
       \frac{{\lambda}}{\sqrt{n}} \sqrt{(\e^{(j)})^\top [\bbK^{(j)}+\lambda I_n]^{-2} \e^{(j)}} + \sigma \sqrt{\frac{\tr[\bbKj]}{4n \lambda}} + \sigma \sqrt{\frac{2}{n}\log(\frac{3}{\delta})}\right\} \\
       &+ \frac{2\sqrt{\tr[\bbKj]}}{n} \left[O(1) + \frac{\sigma}{\sqrt{\lambda}} (\sqrt{n} + \sqrt{2\log(3/\delta)}) \right]+ \sqrt{\frac{\log (3/\delta)}{2n}}.\\
        \end{split}
    \end{equation}
    Under \revise{\Cref{asspt: kernel bound} and \Cref{asspt: tail bound}}, when we take the penalty parameter in the rate $\lambda = O(\sqrt{n})$, we have with high probability that 
    $\| \linearh - f_{0,-j}\| \leq O_p(n^{-1/4})$.
\end{lemma}

\textit{[Proof.]} 
According to \revise{\Cref{lemma: KRR}}, we know that with probability at least $1-\delta/3$,
\begin{equation}
        \|\linearh - f_{0,-j}\|_n \leq   \frac{{\lambda}}{\sqrt{n}} \sqrt{(\e^{(j)})^\top [\bbK^{(j)}+\lambda I_n]^{-2} \e^{(j)}} +
        \sigma \sqrt{\frac{\tr[\bbKj]}{4n \lambda}} + \sigma \sqrt{\frac{2}{n}\log(\frac{3}{\delta})}.
\end{equation}

By \cite{bartlett2002rademacher}, we know that the empirical Rademacher complexity for a function class $\cF_B = \{
f(x) = \alpha^\top \kerf(\X^{(j)}, x): \|f\|_{\cH}\leq B \}$ is bounded as
\begin{equation*}
    \hat{\cR}_{S}(\cF_B) \leq \frac{B\sqrt{\tr[\bbKj]}}{n}.
\end{equation*}

According to \cite{mohri2018foundations}, with probability at least $1-\delta/3$, we have
\begin{equation}
    \begin{split}
        &\sup_{\linearh - h_{\theta_f}\in \cF} \left\{\left\|\linearh(x^{(j)}) - h_{\theta_f}(x^{(j)}) - \left(f_{0,-j}(x^{(j)}) -  h_{\theta_f}(x^{(j)})\right)\right\| - \|\linearh(x) - f_{0,-j}(x^{(j)})\|_n \right\} \\
        \leq& 2\hat{\cR}_S(\cF) + \sqrt{\frac{\log(3/\delta)}{2n}}.
    \end{split}
\end{equation}

From \revise{\Cref{asspt: kernel bound}} and \Cref{lemma: Hnorm_bound}, we have with probability $1-\delta/3$
\begin{equation}
     \| \linearh(x) - h_{\theta_f}(x) \|_{\cH}  := B' \leq O(1) + \frac{\sigma}{\sqrt{\lambda}} \left(\sqrt{n} + \sqrt{2\log(\frac{3}{\delta})}\right).
\end{equation}
Then we have with probability $1-\delta$,
\begin{equation}
    \begin{split}
       & \| \linearh - f_{0,-j}\|\\
       \leq & \|\linearh - f_{0,-j}\|_n + 2\hat{\cR}_S(\cF) + \sqrt{\frac{\log(3/\delta)}{2n}}\\
       \leq & \left\{
        \frac{{\lambda}}{\sqrt{n}} \sqrt{(\e^{(j)})^\top [\bbK^{(j)}+\lambda I_n]^{-2} \e^{(j)}} + \sigma \sqrt{\frac{\tr[\bbKj]}{4n \lambda}} + \sigma \sqrt{\frac{2}{n}\log(\frac{3}{\delta})}\right\} + \frac{2B'\sqrt{\tr[\bbKj]}}{n}+ \sqrt{\frac{\log (3/\delta)}{2n}}\\
       \leq &\left\{
        \frac{{\lambda}}{\sqrt{n}} \sqrt{(\e^{(j)})^\top [\bbK^{(j)}+\lambda I_n]^{-2} \e^{(j)}} + \sigma \sqrt{\frac{\tr[\bbKj]}{4n \lambda}} + \sigma \sqrt{\frac{2}{n}\log(\frac{3}{\delta})}\right\} \\
       &+ \frac{2\sqrt{\tr[\bbKj]}}{n} \left[O_p(1) + \frac{\sigma}{\sqrt{\lambda}} (\sqrt{n} + \sqrt{2\log(3/\delta)}) \right]+ \sqrt{\frac{\log (3/\delta)}{2n}}\\
    \end{split}
\end{equation}

\revise{By the assumptions that $\|[\bbK^{(j)} + \lambda I_n]^{-1}\e^{(j)}\|^2 = O_p(1/\sqrt{n})$ and $\tr[\bbKj] = O_p(n)$ in \Cref{asspt: kernel bound}} (a) (b), when we take $\lambda = O(\sqrt{n})$, we have
\begin{equation}
     \| \linearh - f_{0,-j}\| \leq O_p(n^{-1/4}).
\end{equation}

\subsubsection{Proof of \Cref{lemma: linear_approx}}

\textbf{\Cref{lemma: linear_approx}}
\textit{
For a large neural network whose width is in the order of $O(\sqrt{n})$ where $n$ is the training sample size, our lazy trained neural network is close to its linearization with high probability:
\begin{equation}
    \|\linearh(x) - h_{\theta_f + \Delta \theta_j}(x)\| \leq O(n^{-1/4}).
\end{equation}
}

\textit{[Proof.]}
Since $\linearh(x)= h_{\theta_f} + \Delta \theta_j^\top \nabla_{\theta} h_{\theta}(x)|_{\theta = \theta_f} $ is a linearization of $h_{\theta_f + \Delta\theta_j}(x)$ around the initialization $\theta_f$,
according to Theorem 2.1 in \cite{2020Lee}, when the neural network has a width $M$, the neural network is close to its linearization with probability arbitrarily close to 1:
\begin{equation}
   \| \linearh(x) - h_{\theta_f +\Delta\theta_j}(x) \|_2 = O(\frac{1}{\sqrt{M}}).
\end{equation}
Specifically, when the neural network $M$ takes the order of $O(\sqrt{n})$, we have $\| \linearh(x) - h_{\theta_f +\Delta\theta_j}(x) \|_2 = O(n^{-1/4}).$

\subsubsection{Proof of the Main Theorem (\Cref{thm1}) }

Based on \revise{\Cref{lemma: ridge generalization bound} and \Cref{lemma: linear_approx}} , for a neural network with width at least $M = O(\sqrt{n})$ when the assumptions hold true, by triangular inequality we have
\begin{equation}
    \|h_{\theta_f + \Delta \theta_j} - f_{0,-j}\|\leq \|h_{\theta_f + \Delta \theta_j} -\linearh\| + \|\linearh - f_{0,-j} \| = O_p(n^{-1/4}).
\end{equation}

This holds true for any $j\in[p]$. Then by \cref{lemma: williamson}, we finish the proof for \Cref{thm1}.

\subsubsection{Proof of \Cref{ex:linear}}
\label{proof: example}
The density of $X_1$ given $X_2$ in the setting of \Cref{ex:linear} is:
\begin{equation}
    \begin{split}
        f(x_1|x_2; \rho, \sigma) = \frac{1}{2\pi \sigma^2 \sqrt{1- \rho^2}} \exp{
        \left\{
        - \frac{1}{2(1-\rho^2)\sigma^2}
        (x_1^2 - 2\rho x_1 x_2 + x_2^2)
        \right\}
        },
    \end{split}
\end{equation}
thus we have $X_1| X_2 \sim \cN(\rho X_2, (1-\rho^2)\sigma^2)$.

\section{Additional Experiments}

\subsection{Trace Divergence Rate of the Neural Tangent Kernel Matrix}\label{appendix: trace_div}

\begin{figure}[ht]
    \centering
    \includegraphics[width = 0.6\textwidth]{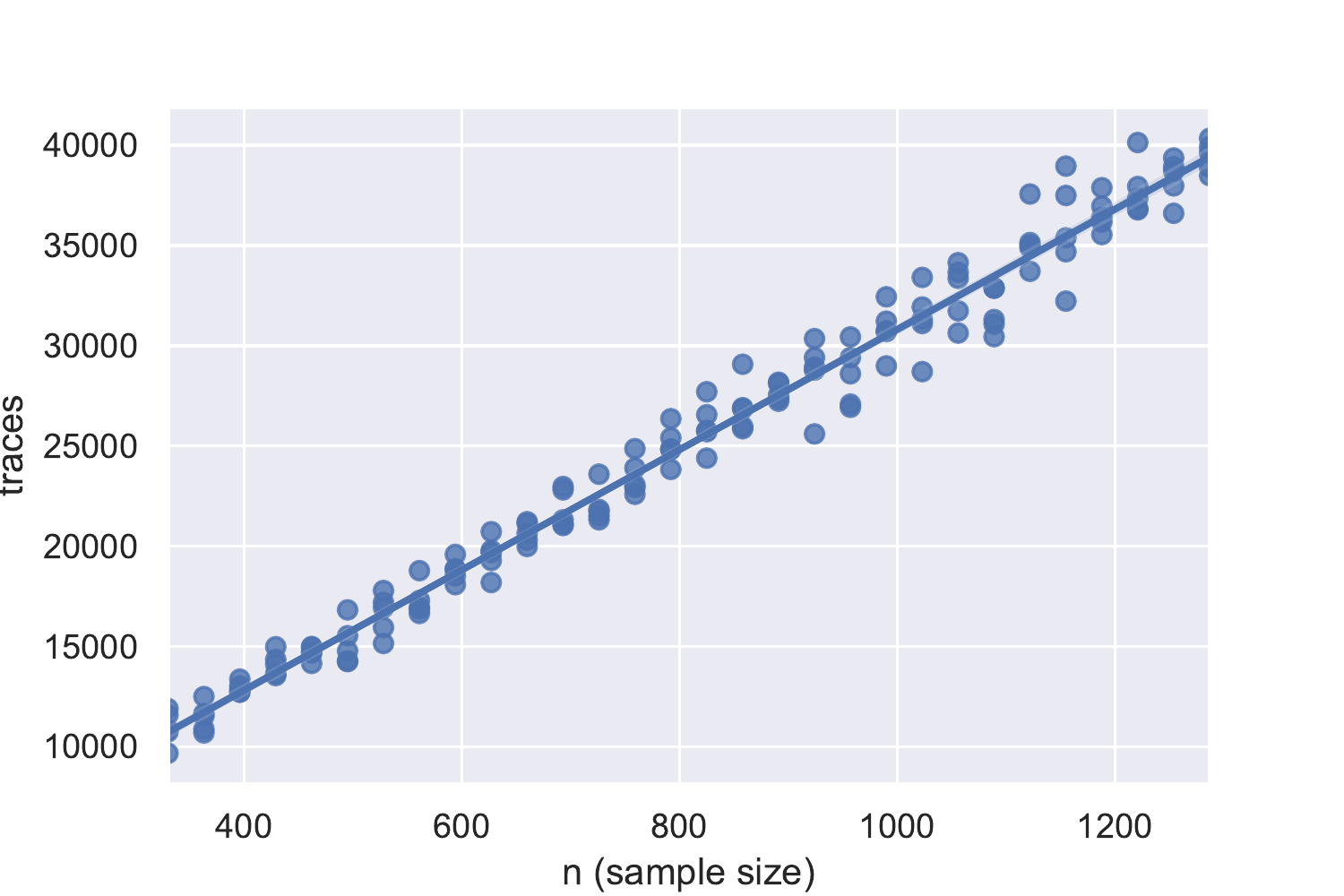}
    \caption{\textbf{The divergence of the NTK matrix trace with respect to the sample size}\newline We calculate the Neural Tangent Kernel(NTK) matrix from a two-layer neural network with $128$ hidden nodes.
    The initialization of the network is the full model parameters trained by the training data, and the input is the test data with the first feature replaced with $0$s. The trace of this NTK matrix is plotted against the sample size of the test data. There is a clear linear pattern in the divergence of the NTK matrix trace w.r.t. the sample size, which verifies Assumption \ref{asspt: trace}.}
    \label{fig: trace_div}
\end{figure}

\revise{
In \Cref{asspt: kernel bound}(b), we assume the trace of the neural tangent kernel matrix with full-model parameters as initialization diverges in the order of $n$ in probability: $\tr(\bbK^{(j)}) = O_p(n)$. In the following experiment, we'll verify this through a simulation.

We consider a two-layer neural network with 128 nodes in the hidden layer. 
The data is generated from a sparse linear model $Y = 1.5 X_1 + 1.2 X_2 + X_3+\epsilon$ where $\epsilon \sim \cN(0, 0.1^2)$ and we have $6$ predictors $X_1, X_2, \dots, X_{6}$ generated from a normal distribution $\cN(0, I_{6} + C)$ with $C_{1,2} = C_{2,1} = 0.5$ and $C_{i,j} = 0$ for $(i,j) \notin \{(1,2), (2, 1)\}$ (this is to add some correlation to the predictors); The total sample size of the data varies in the set $\{1000, 1100,\dots, 4000\}$; Among these simulated samples at each sample size, $2/3$ data are sampled into the training set and $1/3$ samples fall in the testing set. 

We'll first train the full NN model on the training set and get inferred parameters $\hat{\theta}_f$ in the neural network from the full training data. Then we use $\hat{\theta}_f$ as initialization for the reduced neural network. Then we could calculate the neural tangent kernel matrix and its corresponding trace the testing data with one feature (\textit{e.g.} the first feature) dropped and replaced with $0$ (its population mean). We will repeat this process at each sample size level five times (with different random seed to generate data each time), and record the traces with respect to the test data sample size. As shown in \cref{fig: trace_div}, there is a clear trend that the trace diverges linearly as the sample size, which numerically verifies of \Cref{asspt: kernel bound} (b).
}

\subsection{Choosing the Regularization Parameter}\label{appx:cv_alg}
LazyVI involves solving a ridge regression to estimate the difference between the full and reduced model parameters. For variable $j$, we choose the regularization parameter $\lambda_j$ through K-fold cross validation on the prediction made using the estimated $\Delta \theta_j^{\lambda}$. Algorithm \ref{appendix:ridge_param} below shows the entire procedure.

\begin{algorithm}[ht]
\caption{K-Fold CV for $\lambda_j$}\label{appendix:ridge_param}
\begin{algorithmic}
\REQUIRE $\{\X_{i}^{(j)}, Y_i,\e_i^{(j)}, \Phi_i^{(j)} \}_{i=1}^{n_1}$ and $\theta_f$ from Algorithm 1 in main paper; candidate $\lambda$ values $\Lambda$
\STATE Partition $[n_1]$ into $K$ subsets, each denoted $S_t$
\FOR{$\lambda \in \Lambda$}
\FOR{$k = 1, \dots, K$}
\STATE $\Delta \theta_j^\lambda = \arg\min_{\omega\in \bbR^{M}} \frac{1}{n_1 - |S_k|}\sum_{i \notin S_k} 
        [\e_i^{(j)} -  \langle \omega, \Phi_i^{(j)}\rangle ]^2  + \lambda \|\omega\|_2^2$
\STATE $\hat{Y}_i = h_{\theta_f + \Delta \theta_j^\lambda}(\X_{i}^{(j)})$ for $i \in S_k$
\STATE $\epsilon_{\lambda, k} = \frac{1}{|S_k|}\sum_{i\in S_k}(Y_i - \hat{Y}_i)^2 $
\ENDFOR
\STATE $\epsilon_\lambda = \frac{1}{K}\sum_{k=1}^K \epsilon_{\lambda,k}$
\ENDFOR

\ENSURE $\hat{\lambda}_j = \amin_\lambda \{\epsilon_\lambda \}_{\lambda \in \Lambda}$

\end{algorithmic}
\end{algorithm}

\subsection{Full Linear Experiment}
\label{appendix:linear_exp}
\begin{figure}[ht]
\centering
  \includegraphics[scale=0.35]{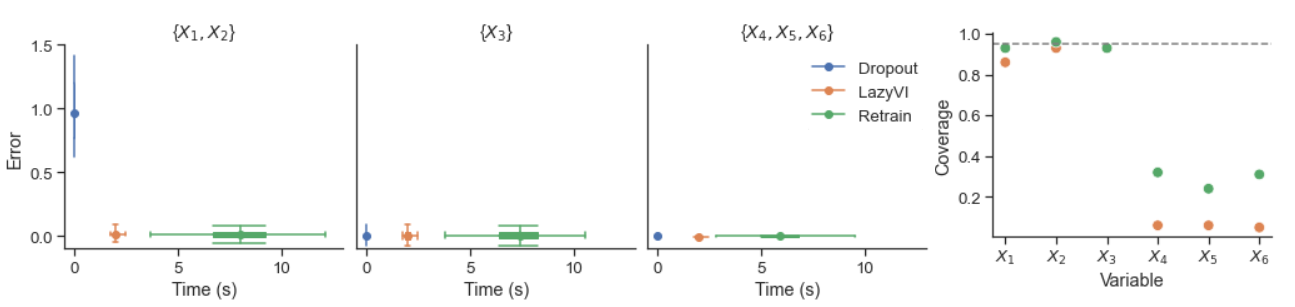}
  \caption{\small \label{fig: time_error}Distribution of computation time vs.\ estimation error relative to retrain ($\hat{\VI} - \VIRT)$ for three different groups of variables: important, correlated $(\{ X_1, X_2\})$; important, uncorrelated ($X_3$); and unimportant, uncorrelated $(\{ X_4, X_5, X_6\})$. 2D box plots show quantiles across 10 repetitions.}
\end{figure}

\Cref{fig: time_error} shows the distribution of computation time vs. VI estimation accuracy for three different groups of variables (important and correlated, important and uncorrelated, unimportant and uncorrelated). We see that LazyVI and retrain are both accurate across all groups of variables, but LazyVI is much faster. Dropout is consistently the fastest method, but is highly inaccurate in estimating $\VI$ for the first group of variables due to their strong correlations. Also in \Cref{fig: time_error} we show the empirical coverage of the LazyVI and retrain 95\% confidence intervals. We see that both retrain and LazyVI achieve desirable coverage for the three important variables; poor coverage of unimportant variables is expected and possibly remedied with a sample-splitting procedure \cite{williamson_unified_2020}.

\subsection{Impact of Lazy Initialization}
\label{appendix:lazy_init}
\begin{figure}[H]
\centering
  \includegraphics[scale=0.5]{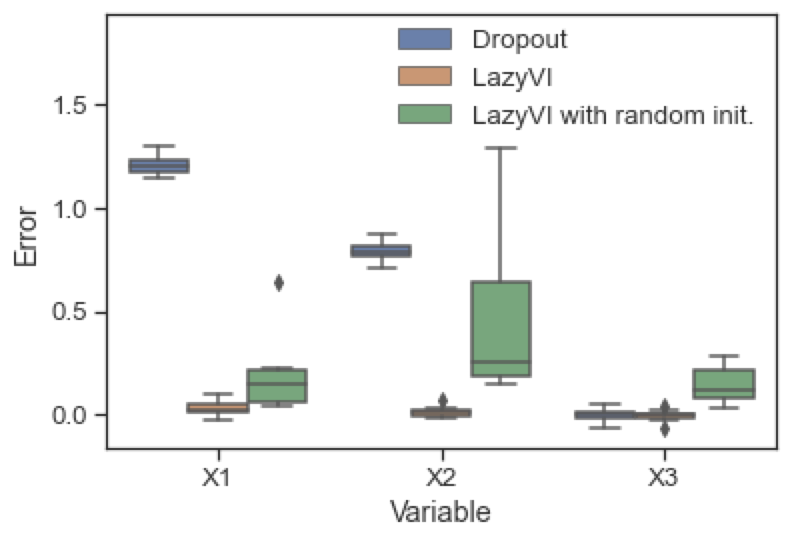}
  \caption{\small \label{fig:lazy_init} Distribution of $\text{VI} - \hat{\text{VI}}$ for the first 3 variables for dropout, LazyVI initialized with the parameters from the full model, and LazyVI with a random initialization.}
\end{figure}

As discussed in the main paper, the initialization of the LazyVI procedure plays a significant role in the accuracy of its estimates. Figure \ref{fig:lazy_init} shows the distribution of the VI error for dropout, LazyVI with a good initialization, and LazyVI with a random initialization across 10 repetitions. We see that the random initialization results in less accurate estimates with high variance.

\subsection{Width of Training Network}
\label{appendix:width}
\begin{figure}[ht]
\centerline{\includegraphics[scale=0.35]{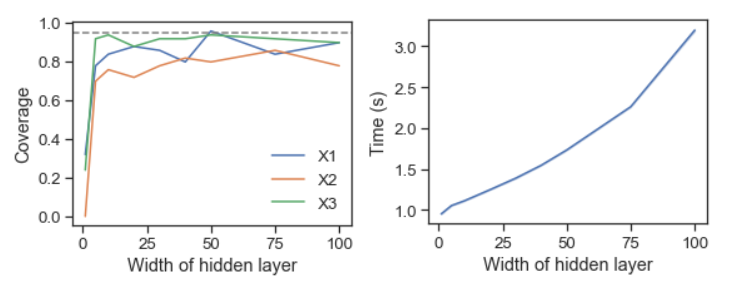}}
\caption{\label{fig:wide}Left: empirical coverage of 95\% confidence intervals (across 50 repetitions) of LazyVI estimates for increasing widths of the training network for the important variables. Dotted line shows 95\% coverage; Right: average computation time for LazyVI with increasing network widths.}
\end{figure}

Theorem \ref{thm1} implies that LazyVI will perform well when the training network is sufficiently wide. \Cref{fig:wide} shows the empirical coverage of the 95\% confidence intervals defined in  \eqref{ci} (across 40 repetitions) for increasing hidden layer widths. We see that coverage increases as the width of the network increases, but the trade-off is that the computation time for LazyVI also increases with the network width (although remains much faster than retraining).

\subsection{Additional Details for Seasonal Forecasting Experiment}\label{appx:climate_details}
For our real data seasonal precipitation forecasting experiment, we use simulations from the Community Earth System Model-Large Ensemble project (CESM-LENS; \cite{kay_community_2015,aws_lens}). CESM-LENS is a 40-member ensemble of climate simulations, where the ensemble members all have the same physics but different initial conditions. From this dataset, we extracted monthly sea surface temperature (SST) records from 1940-2005 on a $1.25^{\circ} \times 0.9^{\circ}$ grid. We compute SST anomalies at each grid point relative to the time period 1950-1989\footnote{\url{https://climatedataguide.ucar.edu/climate-data/nino-sst-indices-nino-12-3-34-4-oni-and-tni}} by subtracting the monthly mean and dividing by the monthly standard deviation, and then we linearly detrend each time series. 

To compute the 10 ocean climate indices (OCI) used in our experiment, we find the average summer (July-October) monthly SST values of these detrended SST anomalies over specified ocean regions. These regions are well established in the literature; we refer to the supplement from \cite{chen_how_2016} to define the boundaries of all OCIs besides NZI, for which we use \cite{mamalakis_new_2018}. See \ref{tab:OCI} for the specific boundaries. As a response, we use the average winter (November-March) precipitation over part of the southwestern US (see \cite{stevens_graph-guided_2021}). We are interested in predicting winter precipitation from the previous summer's SSTs. 

\begin{table}[H]
    \centering
    \begin{tabular}{c|c|c|c|}
        \textbf{Ocean} & \textbf{OCI} & \textbf{Latitude} & \textbf{Longitude}  \\
        \hline
         Pacific & Niño1+2 & $10^{\circ}$S - $0^{\circ}$ & $90^{\circ}$W - $80^{\circ}$W\\
          & Niño3 & $5^{\circ}$S - $5^{\circ}$N & $150^{\circ}$W - $90^{\circ}$W\\
          & Niño3.4 &$5^{\circ}$S - $5^{\circ}$N  & $170^{\circ}$W - $120^{\circ}$W\\ 
          & Niño4 & $5^{\circ}$S - $5^{\circ}$N & $160^{\circ}$E - $150^{\circ}$W \\ 
          & NZI & $40^{\circ}$S - $25^{\circ}$S & $170^{\circ}$E - $160^{\circ}$W \\ 
        \hline 
        Atlantic & TNA & $5^{\circ}$N - $25^{\circ}$N & $55^{\circ}$W - $15^{\circ}$W\\
          & TSA & $20^{\circ}$S- $0^{\circ}$ & $30^{\circ}$W - $10^{\circ}$E\\
                 \hline 
        Indian & SWIO & $32^{\circ}$S - $25^{\circ}$S & $31^{\circ}$E - $45^{\circ}$E\\
          & WTIO & $10^{\circ}$S- $10^{\circ}$N & $50^{\circ}$E - $70^{\circ}$E\\
          & SETIO & $10^{\circ}$S- $0^{\circ}$ & $90^{\circ}$E - $110^{\circ}$E\\
    \end{tabular}
    \caption{Ocean climate indices (OCIs) are defined as the average of the detrended SST anomalies across the regions indicated above. }
    \label{tab:OCI}
\end{table}

Below, we provide an example of SST removal for the experiments in Section \ref{sec:hdcli}. 

\begin{figure}[H]
\centering
  \includegraphics[scale=0.25]{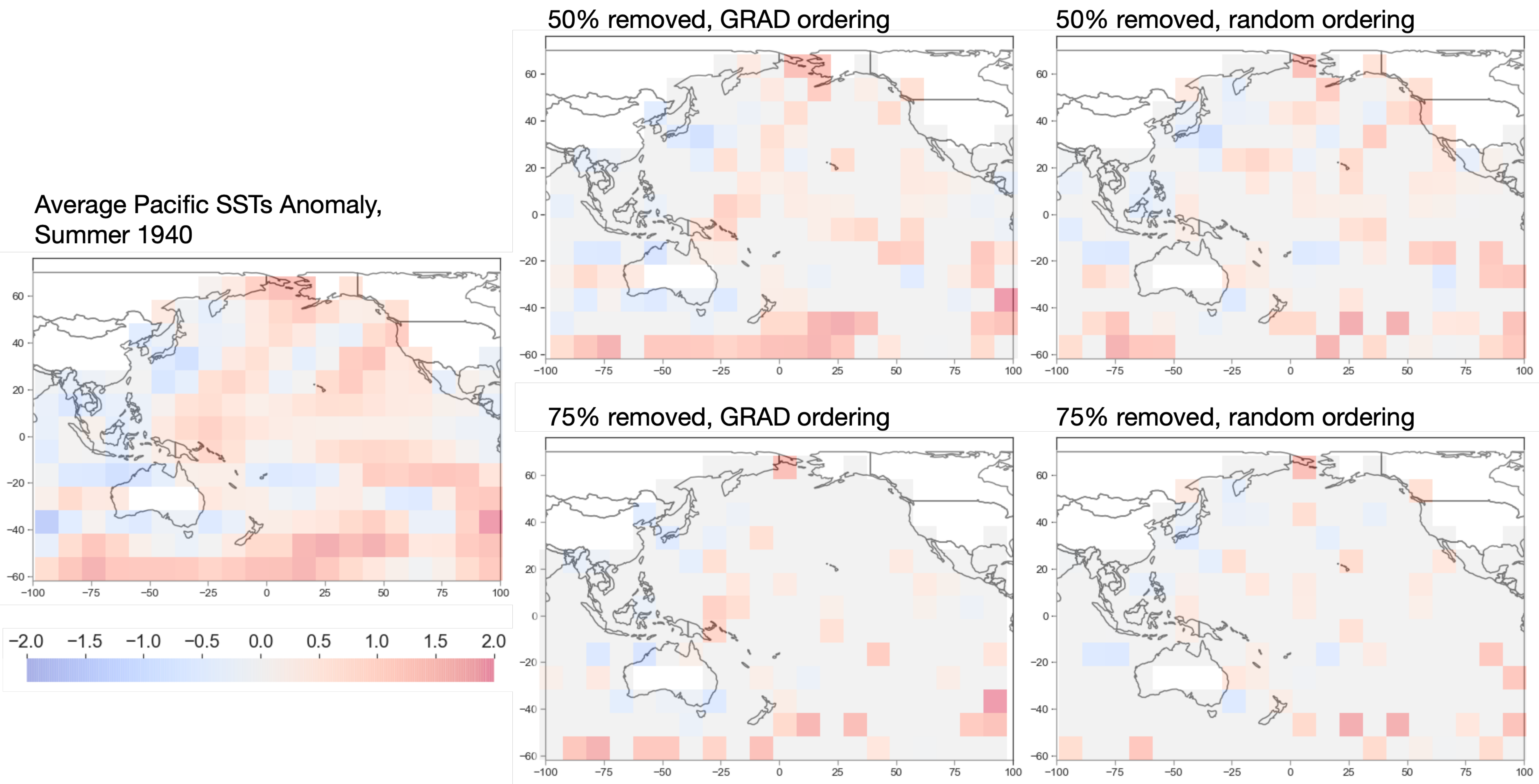}
  \caption{\small \label{fig:saliency} For an example year (1940), Figure \ref{fig:saliency} shows the original image of SST anomalies, and then the modified images after dropping out 50\% and 75\% of the data according to the GRAD importance measure and a baseline random order. In Figure \ref{fig:cli_roar}, we report the decline in predictive performance when removing variables according to these orderings.}
\end{figure}

\subsection{Shapley Value Calculation using Lazy Training Method}\label{appdx: shapley}

As we have discussed in the paper, our method may also provide a faster alternative when calculating Shapley values using large neural networks. We'll indicate this in the following experiment. 

We define population Shapley values in alignment with \cite{williamson_efficient_2020} with an arbitrary measure of predictiveness $V$. For any variable $X_j$, its Shapley value $\psi_j$ is defined as the average gain in oracle predictiveness from including feature $X_j$ over all possible subsets:
\begin{equation}
    \psi_j := \sum_{s\in [p]\setminus \{j\}} \frac{1}{p} {{p-1}\choose |s|}^{-1} \left[ V(f_{0, s\cup \{j\}}, P_0) - V(f_{0, s}, P_0)
    \right],
\end{equation}
where $P_0$ is the true distribution and $f_{0, s}, f_{0, s\cup \{j\} }$ are the oracle prediction functions over the function subset $\cF_s :=\{ f\in \cF: f(u) = f(v) \text{ for all } u,v\in \mathcal{X} \text{ and }u_s = v_s\} $. 
Shapley values have nice properties such as non-negativity, additivity, symmetry, zero for null features, etc. Moreover, such defined Shapley values can assign positive values to collinear variables that are each marginally predictive, whereas previously defined population VIs would assign zero to all collinear variables.

We use the same subset sampling scheme as \cite{williamson_efficient_2020}. However, instead of retraining the model for each subset of features, we generalize our proposed LazyVI method to make the computation more scalable and much faster.  

We are dealing with a Logistic Model with high dimensional sparse features. Specifically, we have $100$ features from a $\cN(0,\Sigma_{100\times 100})$, where the variables are independent except $\text{Corr}(X_1, X_2) = 0.75$. 
The responses are binary, generated from a logistic model: $\log \frac{\bbP(Y = 1)}{1 -\bbP(Y = 1)} = X\beta$, where $\beta = (5,4,3,2,1,0,\dots, 0)^\top \in \bbR^{100}$.

We use a two-layer neural network (with $128$ hidden nodes) and the subset sampling scheme proposed by \cite{williamson_efficient_2020} when calculating Shapley values. We compare the estimated Shapley values and the computing times when we use the retraining method (as is used in \cite{williamson_efficient_2020}) and the lazy training method we proposed.
We calculate the Shapley values on $20$ simulated datasets, each with a sample size $750$. We split each dataset into a training set and test set with sample sizes $500$ and $250$ respectively. We train all the models on the training set, then evaluate the predictiveness loss and calculate the Shapley values on the test set. We set the penalty parameter as $50$ in accordance to the assumption $\lambda = O(\sqrt{n})$.

In the retraining method we'll reconstruct two-layer neural networks with 128 hidden nodes for each subset of features; in the lazy training method however, after training a two-layer neural network with all the features, we use \cref{alg:cap} to train the new model on each subset of features (with all features not included in the subset set equal to their mean). 

\begin{table}[]
\centering
\begin{tabular}{|c|c|c|}
\hline
                           & \textbf{Retrain} & \textbf{Lazy} \\ \hline
\multicolumn{1}{|c|}{Time} & 272.75s          & 50.27s        \\ \hline
Std.                       & 10.8s            & 1.6s          \\ \hline
\end{tabular}
\caption{Average time to compute the Shapley values in one data}
\label{table: shap_time}
\end{table}

\begin{figure}
    \centering
    \includegraphics[width = 0.8\textwidth]{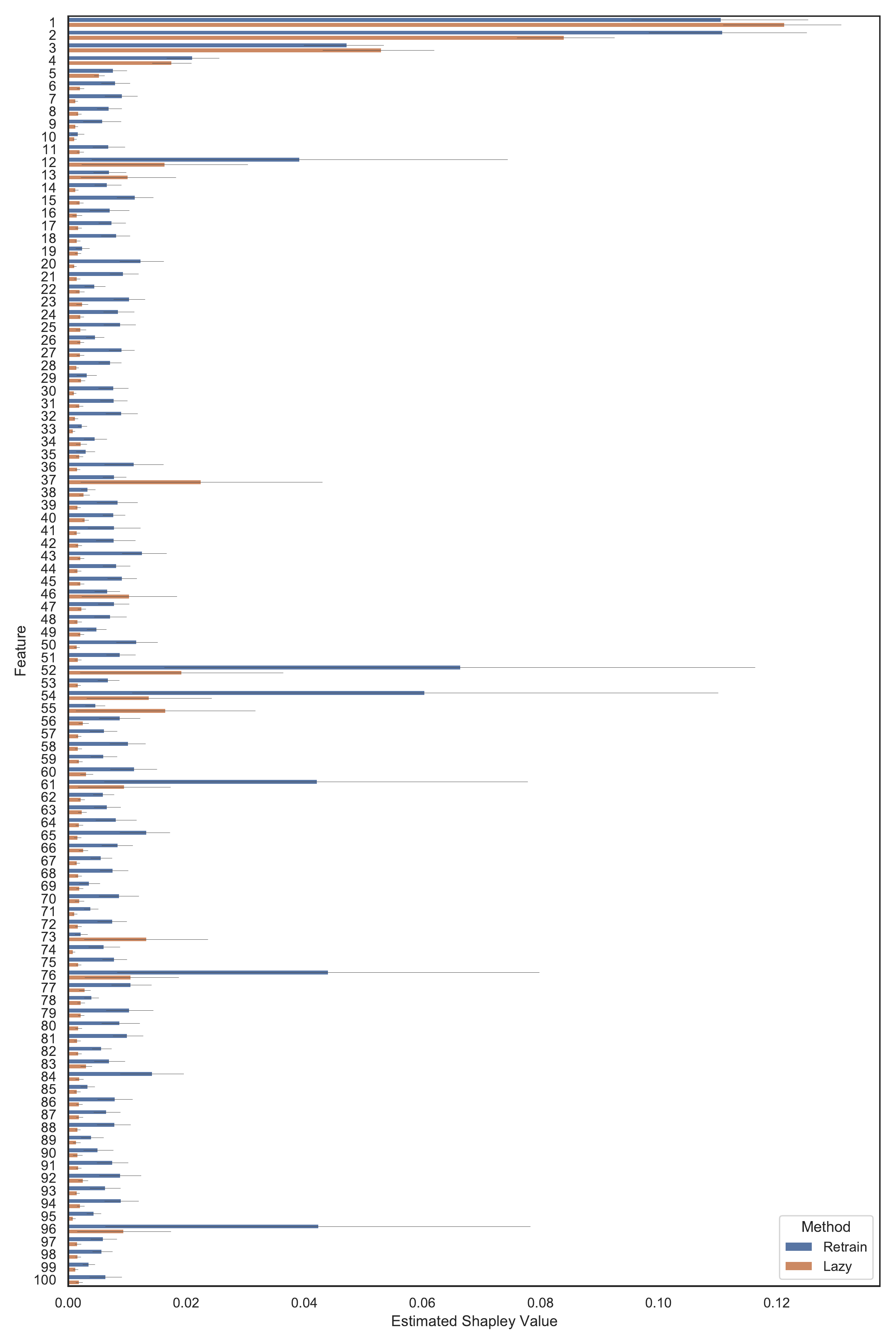}
    \caption{Shapley Values calculated by Retraining vs LazyVI using a two-layer neural network. The data is generated from a logistic model, where only the top 5 features have non-zero weights and all the remaining features have zero weights (The features with zero weights have Shapley values of zero). The experiment is repeated 20 times. The colored bars are the averaged estimated Shapley value of each feature using different methods, and the gray lines indicate the standard deviations. We can see that for non-zero Shapley value variables, the LazyVI estimation is close to Retraining estimation, while for variables with zero Shapley values, retraining estimation has a larger variance than LazyVI.}
    \label{fig: shapley}
\end{figure}

\Cref{table: shap_time} gives the average time to calculate the Shapley values for one data set. Lazy training speeds up the calculation by more than $5$ times. In the meanwhile, we don't sacrifice too much on the Shapley value estimation performance.
As shown in \cref{fig: shapley}, the Shapley value calculated by lazy training is generally close to the retraining results; for those unimportant variables (Feature ID $\geq 6$), Shapley values estimated from the retraining method have a larger variance, as the sample size is relatively small (training size $n = 500$ while $p = 100$), due to the benefit of the regularization step we have in the LazyVI method.

\end{appendices}

\end{document}